\documentclass{article}

\usepackage{microtype}
\usepackage{graphicx}
\usepackage{subcaption}
\usepackage[dvipsnames,table,xcdraw]{xcolor}  %
\usepackage{booktabs} %
\usepackage{listings}
\usepackage{multicol}
\usepackage{enumitem}
\usepackage[most]{tcolorbox}

\usepackage{hyperref}

\usepackage[preprint]{icml2026}

\usepackage{amsmath}
\usepackage{amssymb}
\usepackage{mathtools}
\usepackage{amsthm}

\AtBeginDocument{%
  \setlength{\abovedisplayskip}{2pt plus 1pt minus 1pt}%
  \setlength{\belowdisplayskip}{2pt plus 1pt minus 1pt}%
  \setlength{\abovedisplayshortskip}{0pt plus 1pt}%
  \setlength{\belowdisplayshortskip}{0pt plus 1pt}%
}

\usepackage{tikz}
\usetikzlibrary{positioning}

\usepackage{booktabs}
\usepackage{multirow}
\usepackage{makecell}
\usepackage{tabularx}
\usepackage{array}

\usepackage{natbib}

\usepackage{algorithm}
\usepackage{algorithmic}

\newcolumntype{Y}{>{\centering\arraybackslash}X} %
\newcolumntype{L}[1]{>{\raggedright\arraybackslash}m{#1}} %
\newcolumntype{C}[1]{>{\centering\arraybackslash}m{#1}} %

\newcommand{\best}[1]{\textbf{#1}}

\newcommand{\NA}{\textcolor{black!45}{--}}
\definecolor{oursRow}{RGB}{245,247,255}
\definecolor{blockGray}{RGB}{250,250,250}

\newtheoremstyle{icmlimp}%
{5pt}%
{3pt}%
{\normalfont}%
{0pt}%
{\bfseries}%
{.}%
{0.5em}%
{\thmname{#1}\thmnumber{~#2}\thmnote{:\ \textit{#3}}}%
\theoremstyle{icmlimp}
\newtheorem{implication}{Implication}

\makeatletter
\renewcommand\paragraph{\@startsection{paragraph}{4}{\z@}%
  {1pt}%
  {-0.35em}%
{\normalfont\normalsize\bfseries}}
\makeatother

\usepackage[capitalize,noabbrev]{cleveref}

\newcommand{\eg}{e.g.}

\theoremstyle{plain}

\theoremstyle{definition}

\theoremstyle{remark}

\usepackage[textsize=tiny]{todonotes}

\icmltitlerunning{
  UniLongGen: Taming Long-Horizon Interleaved Image Generation via Context Curation
}

\begin{document}

\twocolumn[
  \icmltitle{
    How Long Can Unified Multimodal Models Generate Images Reliably? \\
    Taming Long-Horizon Interleaved Image Generation via Context Curation
  }

  \icmlsetsymbol{equal}{*}

  \begin{icmlauthorlist}
    \icmlauthor{Haoyu Chen}{hkust}
    \icmlauthor{Qing Liu}{adobe}
    \icmlauthor{Yuqian Zhou}{adobe}
    \icmlauthor{He Zhang}{adobe}
    \icmlauthor{Zhaowen Wang}{adobe}
    \icmlauthor{Mengwei Ren}{adobe}
    \icmlauthor{Jingjing Ren}{hkust}
    \icmlauthor{Xiang Wang}{hust}
    \icmlauthor{Zhe Lin}{adobe}
    \icmlauthor{Lei Zhu}{hkust,hkustcq}
  \end{icmlauthorlist}

  \icmlaffiliation{hkust}{The Hong Kong University of Science and Technology (Guangzhou)}
  \icmlaffiliation{hkustcq}{The Hong Kong University of Science and Technology}
  \icmlaffiliation{adobe}{Adobe Research}
  \icmlaffiliation{hust}{Huazhong University of Science and Technology}

  \icmlcorrespondingauthor{Lei Zhu}{leizhu@hkust-gz.edu.cn}

  \icmlkeywords{Unified multimodal models, long-horizon generation, context management, attention}

  \vskip 0.1in

  \centering
  {
  \includegraphics[width=1\textwidth]{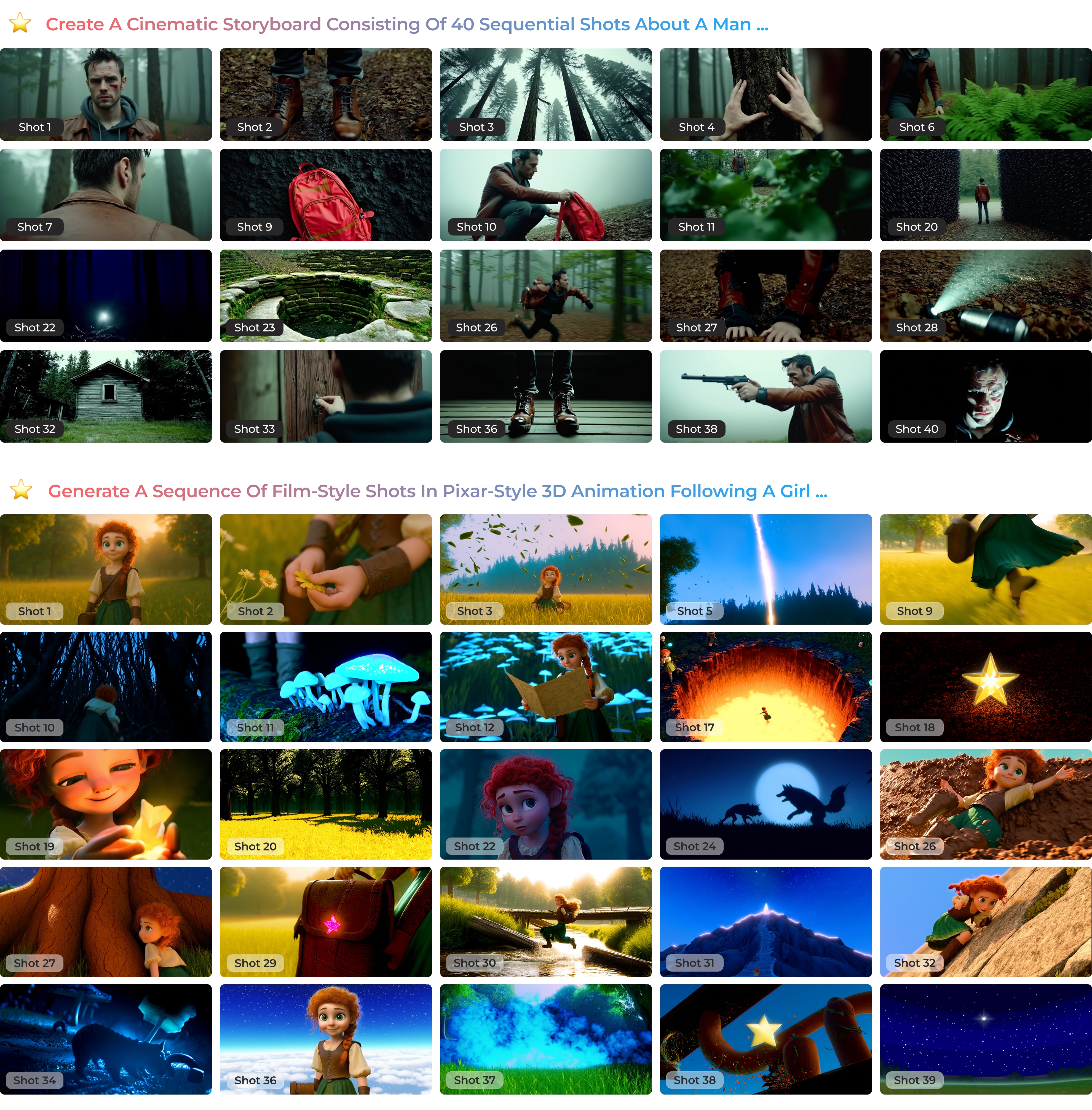}  
  }

  \vskip -0.05in
  {\captionsetup{justification=centering,singlelinecheck=false}%
  \captionof{figure}{\textbf{UniLongGen enables long-horizon interleaved image generation in a single unified sequence.} \newline It generates over 40 images while maintaining high visual quality and cross-image consistency.
  }
  }
  \label{fig:teaser}
  \vskip 0.3in
]

\printAffiliationsAndNotice{}  %

\begin{abstract}
  \vspace{3pt}
  Unified multimodal models hold the promise of generating extensive, interleaved narratives, weaving text and imagery into coherent long-form stories.
  However, current systems suffer from a critical reliability gap: as sequences grow, generation quality rapidly collapses.
  In this work, we investigate the mechanism behind this failure and argue that it is distinct from standard long-context challenges.
  We reveal that in generation, accumulated visual history acts as a source of active pollution, a decay governed specifically by the number of image events rather than raw token count. We identify a structural vulnerability where dense visual tokens overwhelm the attention mechanism, creating "noise" that distorts future synthesis.
  Guided by these mechanistic insights, we propose \textbf{UniLongGen}, a training-free inference strategy that prioritizes safe conditioning over total recall. Instead of retaining all history, UniLongGen dynamically curates the model's memory, identifying and discarding interfering visual signals based on the model's own internal relevance rankings. Extensive experiments demonstrate that this "active forgetting" approach is essential for stability: UniLongGen significantly outperforms baselines in long-horizon fidelity and consistency, while simultaneously reducing memory footprint and inference time.
\end{abstract}

\begin{figure*}[t]
  \centering
  \includegraphics[width=1\textwidth]{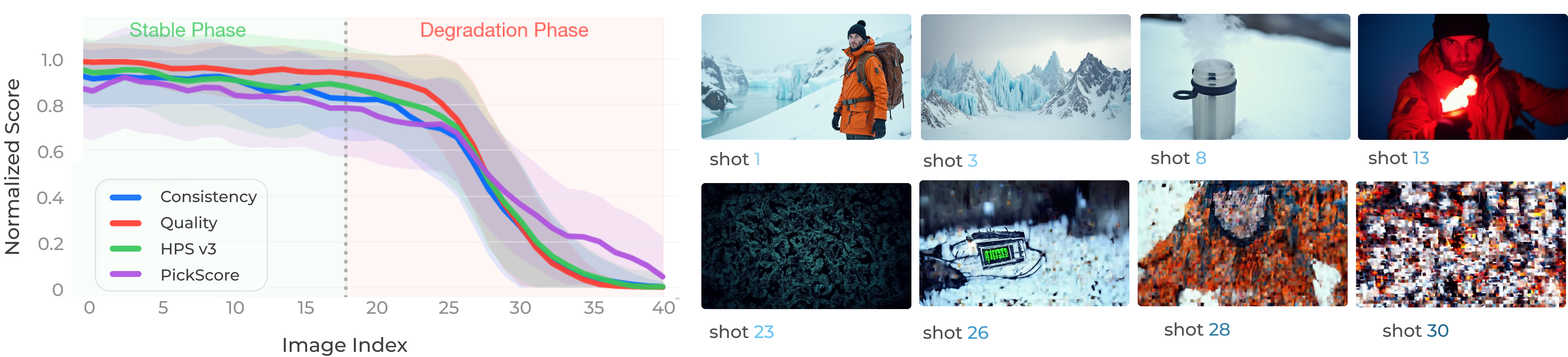}
  \vskip -0.05in
  \caption{\textbf{Generation quality degrades as sequence length increases ($1024{\times}576$).}
    \emph{Left:} Normalized scores for four metrics (Consistency, Quality, HPS v3, PickScore) across a 40-image sequence, averaged over multiple runs (shaded bands indicate variance). All metrics remain relatively stable during the first $\sim$20 images, then undergo a sharp collapse.
    \emph{Right:} Representative samples at different positions. Early shots exhibit high visual fidelity and coherent scene composition, whereas later shots deteriorate into severe artifacts, structural distortions, and ultimately unrecognizable outputs.
  }
  \label{fig:quality_degradation}
  \vskip -0.15in
\end{figure*}

\section{Introduction}

Unified multimodal models (UMMs) are rapidly moving from “describe an image” systems to general-purpose generators that plan, reason, and render in a single autoregressive stream. In this unified setting, a model may write a paragraph, generate an image, continue the narrative, generate the next image, and iterate for dozens of interleaved turns—enabling applications such as character-consistent illustrated stories, long-form storyboarding, and iterative visual design. Yet practitioners repeatedly observe a critical reliability gap: long-horizon interleaved generation collapses. As sequences grow, image quality deteriorates, structure breaks down, and core attributes such as identity and style drift—despite strong performance in short contexts.

Why does this collapse occur? A conventional hypothesis suggests this is merely a manifestation of the "long-context problem," attributable to token overflow or memory constraints. However, we argue that the bottleneck in unified generation is driven by a profound modality asymmetry within the historical context. We observe that in text-dominant sequences, history behaves like an evidence pool: while distant information may fade, preserving more content generally supports better grounding via passive retrieval. In contrast, in vision-heavy interleaved generation, history serves as a dynamic conditioning signal that actively steers the synthesis trajectory. Through a systematic diagnosis, we uncover a structural vulnerability inherent to the unified autoregressive paradigm: \textit{Attention Competition under Dense Visual History}. Unlike sparse text, long-horizon interleaved generation accumulates many previously generated images, each introducing a large set of visual tokens. As these blocks grow, they yield a "heavy-tailed" noise distribution in attention: numerous (often irrelevant) historical visual keys become active competitors, and occasional spurious high-similarity outliers can be exponentially amplified by Softmax to hijack attention, injecting harmful signals that destabilize the current synthesis step.

Our diagnostic analysis yields two primary insights. First, we find that generation quality degrades not as a function of raw token count, but of the number of discrete image events. A model may seamlessly accommodate 100k textual tokens but collapse after fewer than 20 images, suggesting an "Effective Multimodal Context Length" governed by how many distinct visual events the Softmax mechanism can resolve before saturation. Second, we identify that visual history is uniquely detrimental because it can \textit{actively corrupt}, not merely dilute, the conditioning signal. While long text history typically causes passive dilution (leading to vague or weakly grounded outputs), dense visual history induces active pollution: spurious high-similarity matches within historical visual keys capture disproportionate attention mass. This injects incorrect high-frequency details into the current synthesis step, producing structural artifacts and identity/style corruption.
These findings suggest that long-horizon interleaved generation should prioritize safe conditioning over ``remembering everything.'' To maintain fidelity, the model must actively curate its history by suppressing or removing competing historical visual signals that can hijack attention and destabilize synthesis.

Guided by these mechanistic insights, we propose UniLongGen (UNIfied multimodal Context Optimization for interleaved image geneRatioN), a training-free, inference-time context curation policy for long interleaved generation. UniLongGen avoids external retrievers or hand-crafted semantic heuristics, which can be misaligned with the model's generative needs. Instead, it probes the model's own internal relevance signals via a one-shot attention probing pass with dense history, then derives relevance from two complementary depths that reflect layer specialization: (i) an early, text-dependent layer to score historical text blocks for grounding, and (ii) a late, VAE-dominant layer to score historical image blocks for synthesis. We then apply a simple layer-split KV visibility policy throughout generation: early layers attend to the text-filtered history, while late layers attend to the image-filtered history. We implement curation by dropping non-selected tokens directly from the KV cache (rather than compressing them), since compression often preserves spurious ``competitors,'' and can even amplify them, whereas direct dropping more effectively removes the dense visual pollution driving the bottleneck. This reduces softmax competition without introducing distribution-shifting synthetic features.
Our contributions are threefold:
\begin{itemize}
  \item \textbf{Systematic measurement of long-horizon degradation.} We provide reproducible long interleaved generation settings and position-wise degradation curves, introducing effective multimodal context length and the event bottleneck as diagnostic targets.
  \item \textbf{Mechanistic attention diagnosis for unified generation.} We identify interacting mechanisms (dilution, sink tokens, reference erosion) and establish the modality-specific asymmetry between passive dilution (text-heavy history) and active visual pollution (dense visual history), yielding actionable design principles for robust long-horizon synthesis.
  \item \textbf{A training-free context curation policy.} We propose UniLongGen, a plug-and-play inference-time strategy that leverages model-internal attention signals via one-shot probing and enforces a fixed, layer-split KV visibility policy, improving long-horizon fidelity and identity/style consistency while reducing KV-cache footprint.
\end{itemize}

\section{Related Work}
\label{sec:related_work}
\vspace{2pt}

\subsection{Unified Multimodal Models}
\vspace{2pt}

Unified Multimodal Models (UMMs) unify multimodal understanding and generation in a single architecture, broadly following two generative paradigms.
Pure autoregressive (AR) models tokenize all modalities into one sequence for next-token prediction: Chameleon~\cite{chameleon2024} pioneers early-fusion with a unified vocabulary, while Emu3~\cite{emu3_2024} shows next-token prediction alone can be strong. Follow-ups mainly vary visual tokenization, from pixel-level codes~\cite{onecat2025} to high-resolution semantic tokens~\cite{uniworld2025,ovisu1_2025,unitoken2025,tokenflow2024} and learnable query-based representations~\cite{metaquery2025,blip3o2025}.
Hybrid AR-diffusion models pair an AR LLM with a diffusion/flow decoder: Transfusion~\cite{transfusion2024} jointly trains next-token prediction and diffusion; Show-o~\cite{showo2024,showo2_2025} unifies both via omni-attention; JanusFlow~\cite{janusflow2024} combines autoregression with rectified flow. LMFusion~\cite{lmfusion2024} adapts pre-trained LLMs for multimodal generation, and VINO~\cite{vino2026} extends to interleaved omnimodal contexts; other notable models include Hunyuan-Image 3.0~\cite{hunyuanimage3_2025}, NextFlow~\cite{nextflow2026}, OmniGen2~\cite{omnigen2_2025}, and Uni-X~\cite{unix2025}.
To mitigate modality heterogeneity, MoE-style designs such as BAGEL~\cite{bagel2025}, Mogao~\cite{mogao2025}, and HBridge~\cite{hbridge2025} use specialized experts across modalities.

\subsection{Interleaved Generation}

Interleaved image-text generation requires models to produce coherent sequences of arbitrarily alternating modalities. Emu2~\cite{emu2_2023} and SEED-X~\cite{seedx2024} demonstrated strong in-context learning on this task by training on large-scale interleaved corpora such as OBELICS~\cite{obelics2023} and CoMM~\cite{comm2024}. MiniGPT-5~\cite{minigpt5_2023} introduced generative vokens to bridge LLM outputs with visual generation, while DreamLLM~\cite{dreamllm2023} and Kosmos-G~\cite{kosmosg2023} focused on synergistic comprehension-creation learning. Orthus~\cite{orthus2024} employs modality-specific heads for efficient interleaved decoding, and NExT-GPT~\cite{nextgpt2023} generalizes to any-to-any multimodal generation including audio and video.

\section{Quality Collapse in Long Interleaved Sequences}
\label{sec:setup}

\subsection{Problem setting and phenomenon}
\label{sec:problem_setting}

We study long-horizon interleaved generation in unified autoregressive multimodal models: a single contiguous sequence where text segments and discrete image-token blocks are generated alternately, causing the context to grow with each appended image and making later generations condition on an increasingly large, heterogeneous history.
To comprehensively evaluate both long-term memory and dynamic instruction following, we construct diverse narrative scaffolds. These scaffolds are deliberately designed to test global consistency by featuring recurring subjects and persistent styles, while ensuring local diversity through continuous scene transitions, complex actions, and viewpoint shifts. Under this rigorous setting, we generate $N{=}40$ images (interleaved with $T$ text segments) and evaluate, over multiple random seeds, (i) per-image quality and (ii) cross-image consistency as a function of image index $i\in\{1,\dots,N\}$.

We use BAGEL as the base unified multimodal model. BAGEL follows a hybrid autoregressive--diffusion paradigm, combining an autoregressive LLM with a diffusion-based image generator, and represents vision in the same sequence as text to enable interleaved text and image generation. Importantly, it uses both ViT features (semantic) and VAE latents (generation) as visual representations, which aligns with our analysis that separates semantic grounding from synthesis dynamics. 
\Cref{fig:quality_degradation} shows a clear degradation trend as the image index increases: quality metrics drop steadily early on and exhibit a sharp collapse after approximately 20 images. Within a few additional steps, the majority of samples become degenerate and no longer resemble valid images, marking a breakdown of the model's effective generative regime.

\subsection{Observation 1: Image count matters more than token count}
\label{sec:event_bottleneck}

\begin{figure}[t]
  \centering
  \includegraphics[width=\columnwidth]{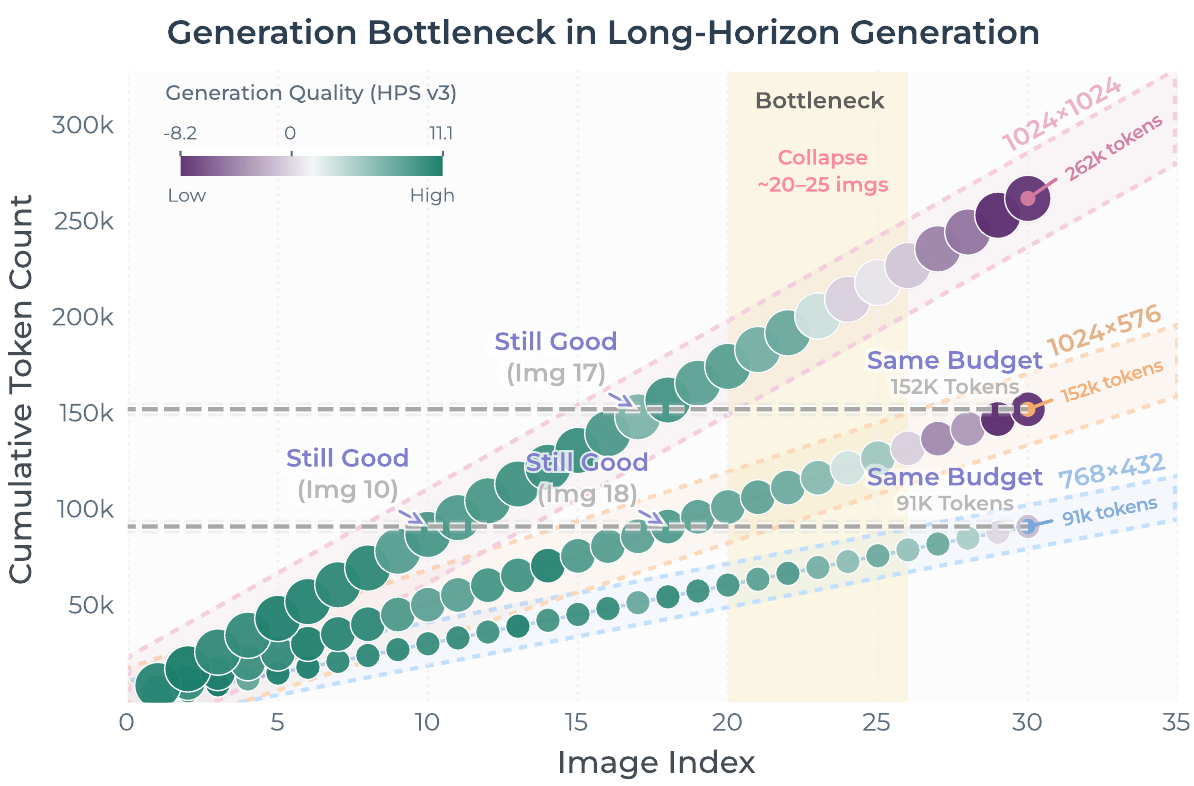}
  \vskip -0.05in
  \caption{\textbf{Bottleneck: Generation quality depends on image count, not token length.}
    We visualize generation quality (HPS v3, color scale) for long-horizon sequences with different tokens-per-image rates ($1024{\times}1024$, $1024{\times}576$, $768{\times}432$).
    Despite wide disparities in cumulative token count (y-axis), all settings exhibit quality collapse within the same image-index window (20--25 images, shaded region).
  Horizontal comparisons (dashed gray lines) confirm that matching the token budget does not predict success; instead, the number of semantic events (image index) is the dominant bottleneck.}
  \label{fig:resolution_invariance}
  \vskip -0.2in
\end{figure}

We first investigate whether this degradation is driven by the sheer number of tokens (a computational or memory capacity limit) or by the number of distinct semantic events (a semantic capacity limit). We compare long-horizon runs across three image resolutions, which create vastly different token growth rates.
Crucially, as shown in \cref{fig:resolution_invariance}, we find that quality collapses within the same \emph{semantic window} of approximately 20--25 images across all settings, regardless of the total token count. This is most evident when controlling for memory budget: at a fixed depth of $\sim$150k tokens, a 17-image sequence ($1024{\times}1024$) maintains high fidelity, whereas a 30-image sequence ($1024{\times}576$) has completely degraded. This implies an \emph{effective multimodal context length} governed by the number of distinct images or events that attention can reliably use for conditioning, rather than raw sequence length.

It rules out simple explanations based on computational limits. The model can handle 150k tokens when they represent fewer images, but fails at the same token count when they represent more images. The degradation is therefore not a simple matter of ``running out of memory,'' but a structural consequence of accumulating visual history. We term this the \textbf{Event Bottleneck}: the effective context length is limited by the number of visual events, not the number of tokens.
This finding shifts the optimization goal from \emph{fitting more tokens} (via compression or architecture) to \emph{managing event-level competition} (via curation). To understand why visual history is so problematic, we next compare it against text-only history of equivalent length.

\begin{figure}[t]
  \centering
  \includegraphics[width=\columnwidth]{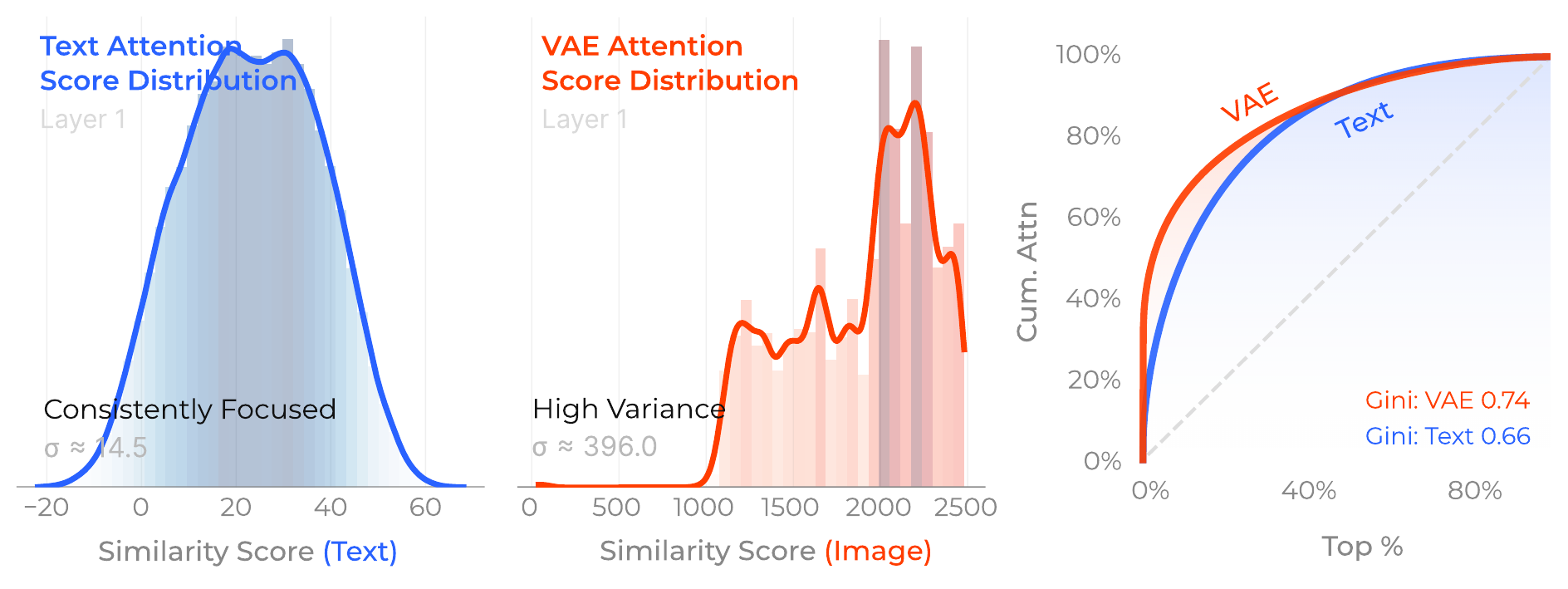}
  \includegraphics[width=\columnwidth]{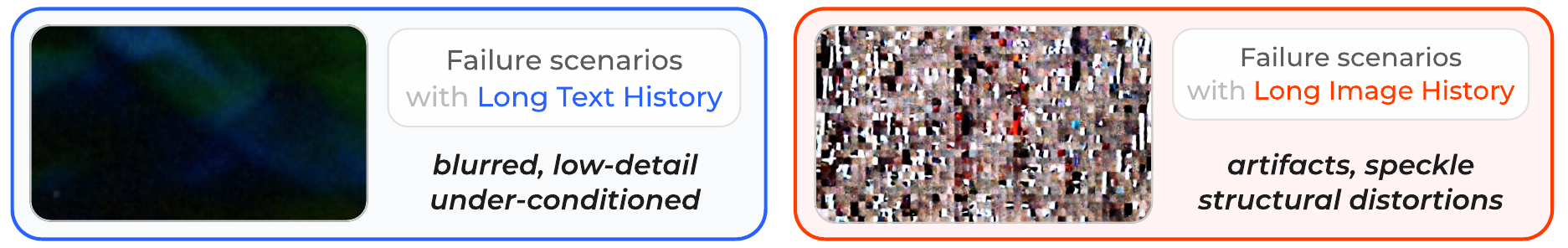}
  \vskip -0.05in
  \caption{\textbf{Token-matched text vs.\ image history induces distinct attention regimes and failure modes.}
    We compare (A) a 120K-token text-only history and (B) a token-matched image-heavy history (22 images).
    \textbf{Top row (diagnostics):} similarity-score distributions from current-image queries to historical tokens show that text history produces a consistently concentrated score spectrum, while image-heavy history exhibits much higher variance and a heavier tail (outliers).
    A concentration plot (cumulative attention vs.\ top-\% tokens) further indicates that attention under image history is more top-heavy (higher Gini).
    \textbf{Bottom row (outcomes):} token-matched text history mainly leads to \emph{passive dilution} (blurred, under-conditioned generations), whereas token-matched image history causes \emph{active pollution} (artifacts, speckle, and structural distortions).
  }
  \label{fig:hybrid_analysis}
  \vskip -0.15in
\end{figure}

\subsection{Observation 2: Modality-Asymmetric Failure under Long History}
\label{sec:gen_not_understanding}

The Event Bottleneck points to visual history as the main driver of long-horizon collapse. A natural follow-up is whether this collapse is simply another instance of long-context failure—i.e., the same phenomenon seen in long-context understanding—or whether unified generation exhibits a qualitatively different failure mode.

\paragraph{Text history dilutes; image history pollutes.}
We design a controlled comparison matching total context length while varying history type: (A) a 120K-token text-only history versus (B) a token-matched image-heavy history containing 22 images. If degradation were primarily a function of raw context length, these two settings should produce similar failure modes.
\cref{fig:hybrid_analysis} shows a clear qualitative gap. With long text-only history, failures are mostly passive: generations become generic or under-conditioned (\eg, blurry details, weaker adherence to the prompt), consistent with retrieval inefficiency where relevant evidence is missed or under-weighted. In contrast, with token-matched image-heavy history, failures are active and structural: artifacts, distortions, and identity/style drift appear, suggesting that spurious visual competitors inject incorrect signals into the conditioning used for synthesis.

\paragraph{The unique risk of dense visual history.}
This asymmetry highlights a fundamental distinction in how different modalities affect the generation process. With text-dominant history, failures resemble standard long-context retrieval issues: irrelevant text primarily harms performance by diluting attention, leading to a passive failure of under-conditioning. In contrast, dense visual history introduces a structural vulnerability. Because historical visual tokens share the same representational space as the target image, they actively steer the pixel-level synthesis trajectory. Consequently, irrelevant visual features are not just passively missed; they are mixed into the conditioning representation and compounded across steps. This confirms that long-horizon collapse is not a ``context length'' issue, but a modality-specific vulnerability where dense visual keys trigger active pollution.

This finding squarely motivates the next section's deep dive into the underlying mechanisms, where we investigate precisely how visual history creates spurious attention competition to hijack the generation process.

\vspace{-2pt}
\section{Mechanism: Attention Hijacking via Visual Pollution}
\label{sec:diagnosis}

The previous section established \emph{what} goes wrong: visual history causes active pollution, not mere dilution.

This section explains \emph{why} this happens in unified generation. We first frame long-horizon failure as a zero-sum \emph{allocation} problem under the softmax attention budget, and provide evidence via macro-level entropy growth and micro-level key-reference erosion (\cref{sec:attn_misalloc}). We then introduce the core mechanism of \emph{tail-risk hijacking} under dense visual keys, showing how heavy-tailed outliers amplified by softmax turn dilution into corruption (\cref{sec:tail_risk}). Finally, we highlight an important nuance for solution design: attention dynamics are specialized across synthesis steps and transformer depth, motivating depth-aware policies rather than a single uniform relevance mask (\cref{sec:step_layer_specialization}).

\begin{figure}[t]
  \centering
  \includegraphics[width=1\columnwidth]{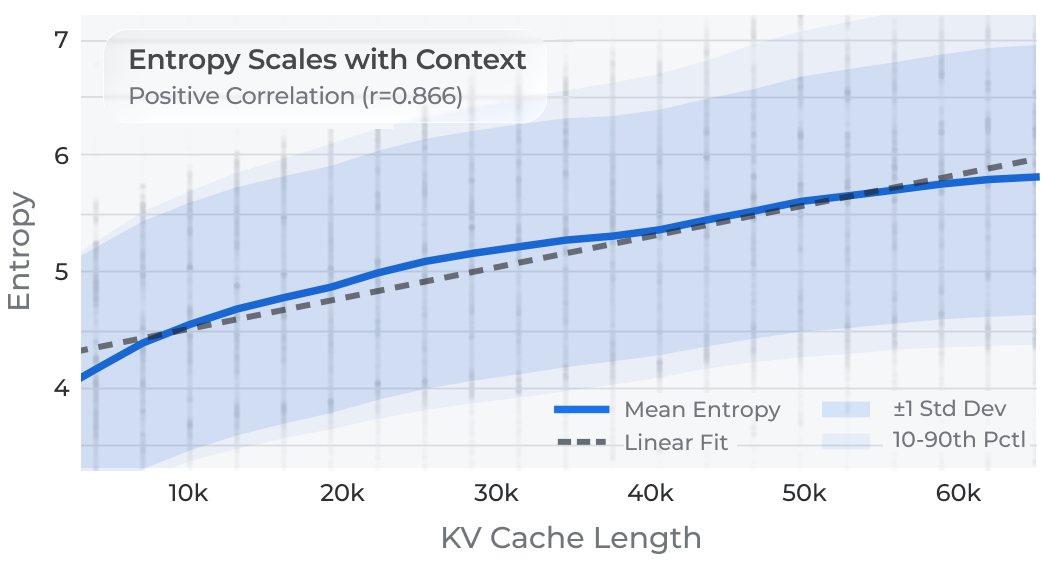}
  \vskip -0.05in
  \caption{\textbf{Attention becomes diluted and unfocused as context grows.}
  Attention entropy rises with context length, indicating the model becomes increasingly ``confused'' about where to attend.}
  \label{fig:attn_entropy}
  \vskip -0.05in
\end{figure}

\begin{figure}[t]
  \centering
  \includegraphics[width=\columnwidth]{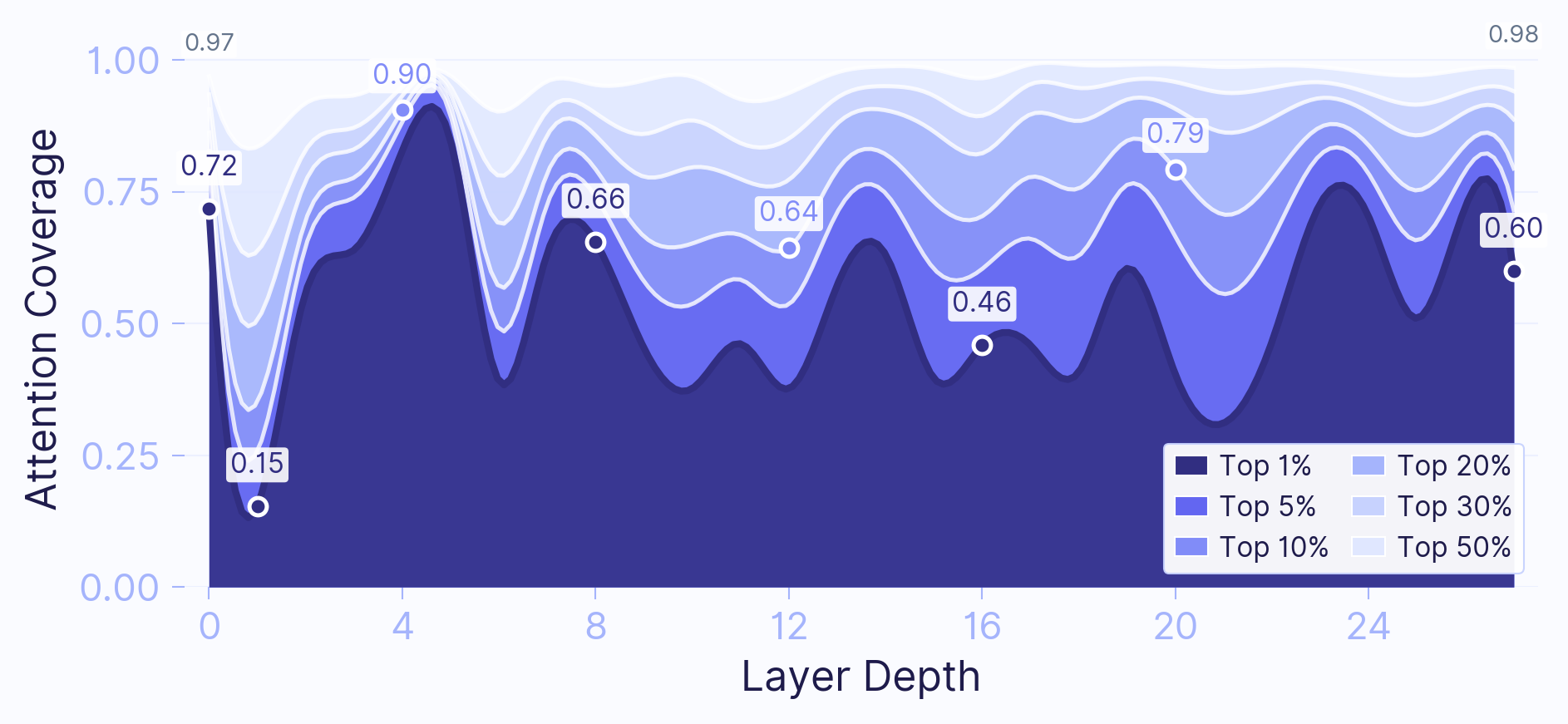}
  \vskip -0.05in
  \caption{\textbf{Attention is inherently top-heavy across depth.}
    We measure concentration as the cumulative attention \emph{coverage} captured by the top-$k\%$ tokens in each layer (shaded bands).
  }
  \label{fig:entropy_coverage_tradeoff}
  \vskip -0.2in
\end{figure}

\begin{figure*}[t]
  \centering
  \begin{minipage}[t]{0.63\textwidth}
    \centering
    \includegraphics[width=\linewidth]{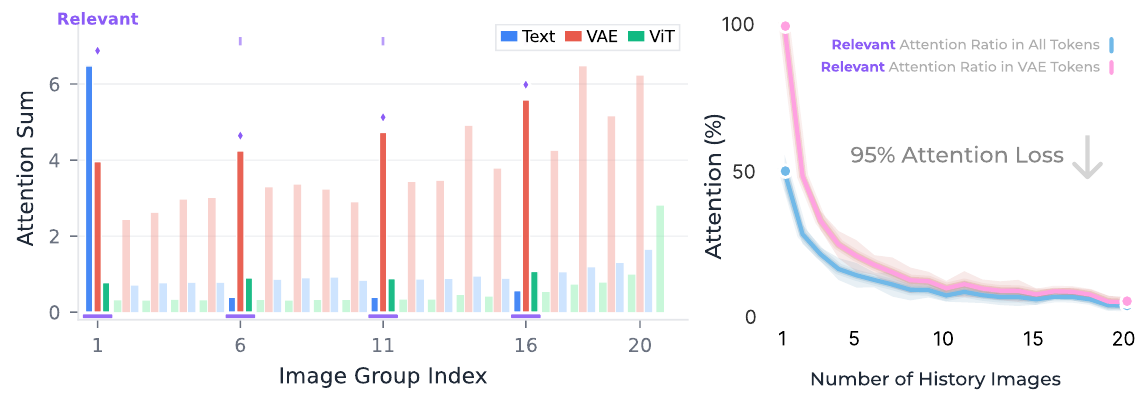}
    \vskip -0.05in
    \captionof{figure}{\textbf{Attention misallocation under redundant image history.}
      \emph{Left: Irrelevant history still attracts attention.} We aggregate attention over historical image groups (by token type/modality) while generating a target character; only a few groups (e.g., Image$_{1,6,11,16}$) contain character-relevant references.
      Yet many irrelevant images receive non-trivial attention, diluting the budget for true references.
      \emph{Right: Redundancy rapidly attenuates key-reference attention.} In a controlled ``one-key-reference + $N$ distractors'' setup, the key-reference attention share drops steeply as distractors are appended (e.g., $\sim$50\% $\rightarrow$ $\sim$4\%, a $\sim$95\% relative loss).
    }
    \label{fig:visual_memory}
  \end{minipage}\hfill
  \begin{minipage}[t]{0.35\textwidth}
    \centering
    \includegraphics[width=\linewidth]{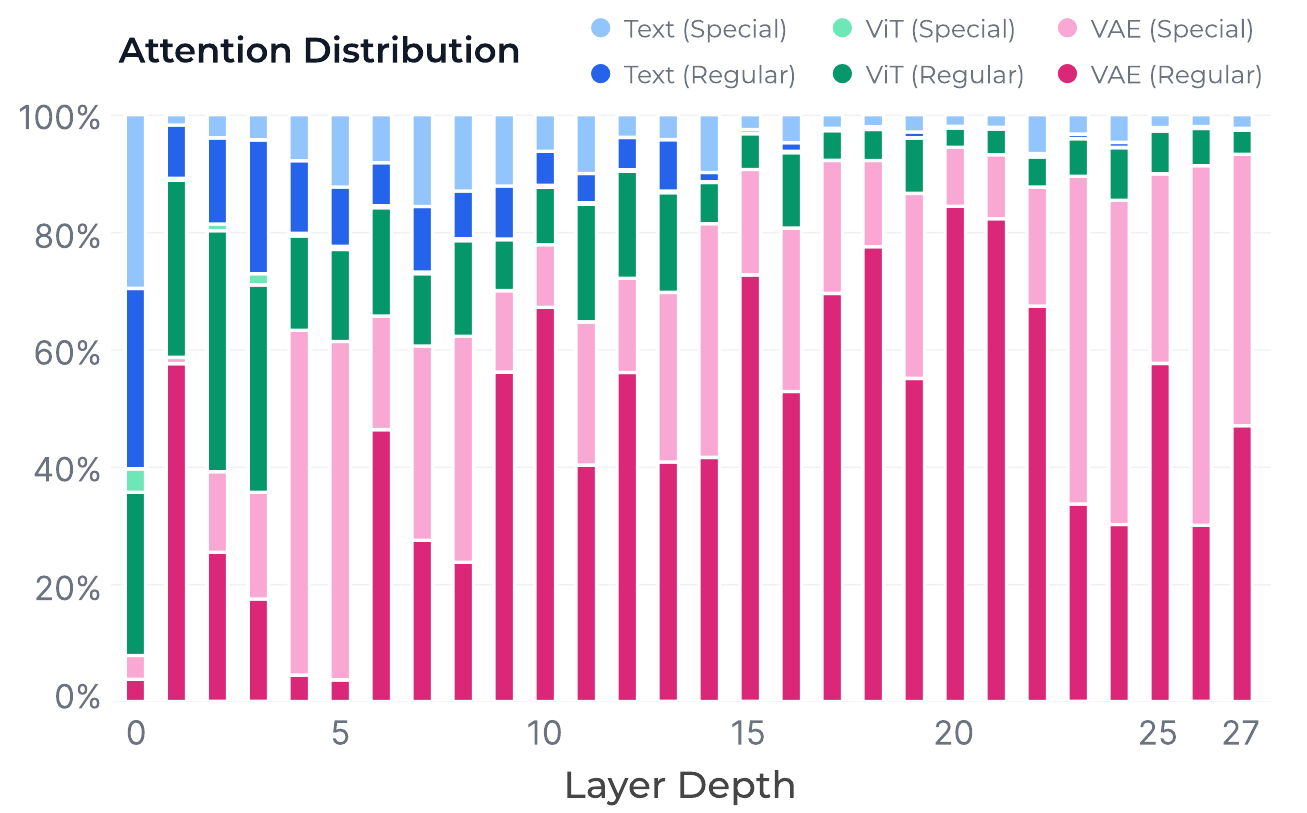}
    \vskip -0.05in
    \captionof{figure}{\textbf{Depth-wise modality specialization in attention.}
    Layer-wise attention ratios show a clear shift in where the model allocates its budget: earlier layers attend more to Text/ViT tokens (grounding and multimodal routing), while later layers increasingly prioritize VAE tokens that drive image synthesis. Special/control tokens remain a persistent share of attention across depth.}
    \label{fig:token_type_ratio}
  \end{minipage}
  \vskip -0.1in
\end{figure*}

\subsection{Visual Competition and Attention Budget Collapse}
\label{sec:attn_misalloc}

To explain the degradation observed in \cref{sec:event_bottleneck,sec:gen_not_understanding}, we view long-horizon failures as an \emph{allocation} problem under a zero-sum softmax budget: as the context accumulates more image events, attention must be distributed across an expanding set of competitors, making it increasingly difficult to reliably \emph{resolve} the few conditioning cues that actually matter for synthesis.

\paragraph{Attention Entropy Grows with Context.}
We first quantify how focused the model's attention is during image generation using the \emph{entropy} of the attention distribution over context tokens.
High entropy indicates a more diffuse distribution, whereas low entropy indicates a concentrated focus.
We observe a systematic increase in attention entropy as the context grows (\cref{fig:attn_entropy}), especially when additional \emph{image} tokens are present.
This macro signal suggests rising uncertainty: with more competing tokens, attention becomes less consistently anchored to a small, stable set of cues.

\paragraph{Competing History Erodes Key References.}
To directly measure whether the model can still retrieve the \emph{right} conditioning signal, we track the \emph{key-reference attention mass}:
the attention allocated to tokens of a known relevant historical image, normalized by total historical-image attention.
As distractors accumulate, the key-reference share drops sharply (\cref{fig:visual_memory}), while visual quality degrades with position (\cref{fig:quality_degradation}).
Together, these curves show that the relevant signal remains in-context, yet becomes progressively harder to resolve; degradation is thus primarily an \emph{allocation failure}, not an \emph{information failure}.

\subsection{Active Pollution from Tail-Risk Hijacking}
\label{sec:tail_risk}

\paragraph{Phenomenon: image history induces heavy-tailed matches and top-heavy attention.}
Compared to token-matched text history, image-heavy history exhibits a markedly heavier-tailed similarity-score distribution (\cref{fig:hybrid_analysis}, right): a single image contributes thousands of patch-level keys, so even semantically irrelevant images have a non-trivial chance of producing high dot-product \emph{outliers} with the current query (incidental alignments).
Unified attention is also inherently top-heavy (\cref{fig:entropy_coverage_tradeoff}), with the top-$10\%$ tokens often capturing over half of the budget; consequently, a small number of extreme visual matches can dominate attention.
This explains why degradation is governed by \emph{events} rather than raw token count (\cref{sec:event_bottleneck}): each appended image introduces a fresh pool of dense keys, increasing the probability of encountering an outlier that captures the budget.

\paragraph{Softmax amplifies tail risk into hijacking, turning dilution into corruption.}
Because Softmax acts as an exponential amplifier, a spurious similarity outlier does not merely add background noise---it can \emph{hijack} a massive share of the attention budget. By locking onto these irrelevant high-frequency patches, the model overrides its intended conditioning (\eg, true subject references or prompt cues). Consequently, it directly \emph{injects} incorrect textures and structural artifacts into the current synthesis step, turning what would normally be passive context dilution into active visual corruption.
This modality-asymmetric failure mode is therefore not a generic 
long-context retrieval issue, but a tail-risk phenomenon induced by 
dense visual keys and amplified by Softmax.

\begin{implication}[Event-level curation is required to eliminate tail risk]
  Tail-risk hijacking implies that long-context techniques designed to \emph{fit} more tokens (e.g., token compression or generic sparse attention) are insufficient for unified generation: they may retain the very high-variance outliers that trigger pollution.
  Stability requires removing the source of competition, motivating \textbf{event-level curation} (discarding entire image blocks based on relevance) to truncate the heavy tail and prevent hijacking.
\end{implication}

\begin{figure}[t]
  \centering
  \includegraphics[width=1\columnwidth]{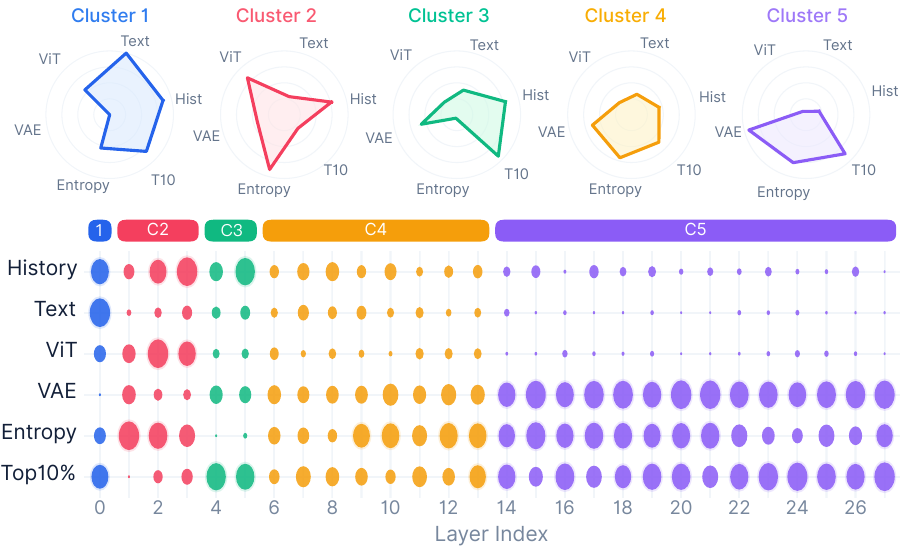}
  \vskip -0.05in
  \caption{\textbf{Layer-wise functional clustering.}
    Ward hierarchical clustering on six attention features reveals five functional regimes.
    \textcolor[RGB]{66,133,244}{\textbf{C1}} (Layer~0) is text-dominant, \textcolor[RGB]{234,67,101}{\textbf{C2}}--\textcolor[RGB]{52,168,133}{\textbf{C3}} balance text/vision, and \textcolor[RGB]{251,176,59}{\textbf{C4}}--\textcolor[RGB]{136,106,234}{\textbf{C5}} (late layers) are VAE-dominant with higher entropy; \textbf{Bottom:} bubble size indicates feature magnitude, and contiguous assignments show a structured shift from grounding to synthesis.
  }
  \label{fig:layer_clustering}
  \vskip -0.3in
\end{figure}

\begin{figure*}[t]
  \begin{center}
    \centerline{\includegraphics[width=\linewidth]{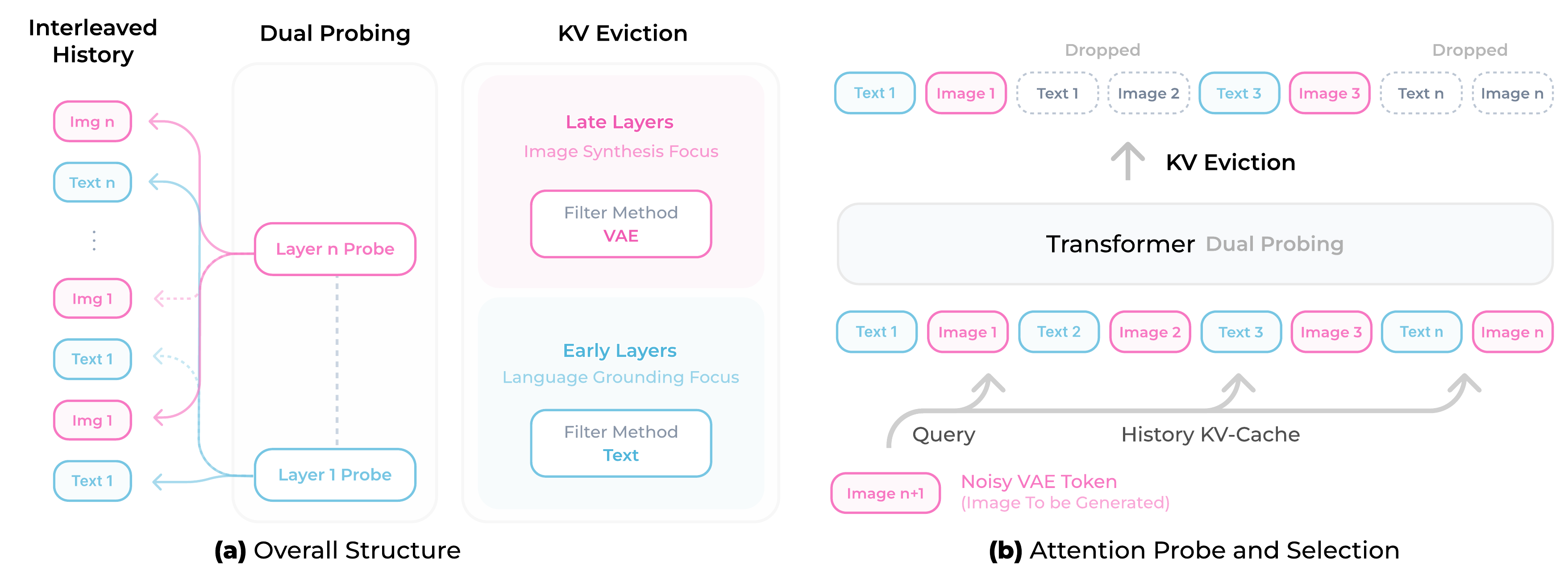}}
    \vskip -0.05in
    \caption{
      \textbf{Overview of UniLongGen.}
      Given a long interleaved history, we perform a one-shot probing pass with dense KV and derive two relevance masks from model-internal attention:
      (i) at an early, text-dependent layer (Layer 1), we score \emph{historical text blocks} using current VAE queries with a length-normalized Pre-Softmax similarity score;
      (ii) at a late, generation-dominant layer (default: $\ell_{\text{syn}}{=}15$), we score \emph{historical VAE image blocks}.
      We then keep the masks fixed and apply a layer-split KV visibility throughout generation: early layers use the text mask, while late layers use the image mask, dropping non-selected tokens.
    }
    \label{fig:method_overview}
  \end{center}
  \vskip -0.2in
\end{figure*}

\vspace{-2pt}
\subsection{Step- and layer-wise specialization}
\label{sec:step_layer_specialization}

Unified generators built on diffusion / rectified-flow refine the same image tokens over multiple steps.
Across steps, we observe a systematic reallocation of attention between historical context and current-generation tokens (\cref{fig:stepwise_attn}).

Transformer depth is specialized in unified generation.
Early layers allocate more attention to text and multimodal routing, while late layers become increasingly VAE-dominant and directly shape the synthesis trajectory (\cref{fig:token_type_ratio}).
Layer clustering on attention features further reveals contiguous functional blocks, separating a text-dependent early regime from a generation-dominant late regime (\cref{fig:layer_clustering}).

\Cref{fig:token_type_ratio} breaks down attention by modality (Text/ViT/VAE) and by token type (regular vs.\ special).
Across layers, a non-trivial fraction of attention mass is consistently allocated to \emph{special} tokens (e.g., \texttt{<bos>}, modality separators, and interleaving delimiters) within each modality, rather than to regular content tokens alone.
These structural tokens provide stable markers for segmentation and modality/state transitions in unified generation, and thus attract attention as the network composes multimodal information.

\begin{implication}[layer-split relevance estimation]
  The functional shift from text-dominant grounding (early layers) to VAE-dominant synthesis (late layers) implies that a single relevance mask is insufficient. This motivates UniLongGen's dual-probe design: we use text-based relevance for early layers to ensure instruction following, and VAE-based relevance for late layers to maintain consistency.
\end{implication}

\vspace{-5pt}
\section{UniLongGen: Self-Guided Context Curation to Prevent Visual Attention Hijacking}
\label{sec:method}

Long-horizon interleaved generation fails because accumulated visual history can actively pollute synthesis by hijacking attention (\cref{sec:diagnosis}).
UniLongGen targets this setting and is applied only during image-synthesis steps: instead of fitting or compressing the full history, we enforce relevance-filtered sparsity by retaining a small set of historical reference images and evicting the rest from the KV cache to reduce softmax competition.

\textbf{Overview.}
For each new image, UniLongGen executes a three-step pipeline (\cref{fig:method_overview}):
(i) \textbf{One-Pass Context Profiling:} Perform a single forward pass with the full history (dense KV) to probe model-internal attention.
(ii) \textbf{Dual-Depth Scoring:} Derive relevance scores for historical text blocks and VAE image blocks from two specialized network depths.
(iii) \textbf{Layer-Split Generation:} Generate the image using a fixed, layer-split KV-visibility policy that keeps only the selected history.
This decouples relevance estimation from generation and enables a stable context policy across all subsequent diffusion/flow steps.

\subsection{Attention-Based Relevance Scoring}
To robustly identify relevant history, we propose a pre-softmax aggregation strategy that reduces the noise from spurious high-frequency matches in individual heads.
We interpret the relevance of a historical block $B$ (representing either a text segment $T$ or a VAE image $V$) as the expected similarity between its keys and the average representation of the current image's queries.
We write the interleaved history as $m$ turns indexed by $i\in\{1,\dots,m\}$.
Each turn $i$ contains a text block $T_i$ and a VAE image block $V_i$, where each block denotes the set of token indices belonging to that segment in the serialized sequence (including boundary special tokens).
Let $V^{\mathrm{cur}}$ denote the set of VAE tokens for the current image being synthesized.
Let $\mathbf{q}_{v}^{(\ell)}$ and $\mathbf{k}_{u}^{(\ell)}$ denote the multi-head representations (shape $H{\times}d$) for a current query token $v$ and a historical key token $u$ at layer $\ell$.
We use $\mathbf{q}_{v,h}^{(\ell)}\in\mathbb{R}^d$ and $\mathbf{k}_{u,h}^{(\ell)}\in\mathbb{R}^d$ to denote the $h$-th head slice.
We first condense the current image's generation intent into a single mean query vector $\bar{\mathbf{q}}^{(\ell)}$ by averaging over all spatial VAE tokens:
\begin{equation}
  \bar{\mathbf{q}}^{(\ell)} \triangleq \frac{1}{|V^{\mathrm{cur}}|} \sum_{v \in V^{\mathrm{cur}}} \mathbf{q}_{v}^{(\ell)} .
\end{equation}
We then define the relevance score $S(B; \ell)$ for any historical block $B$ (text or VAE) as the average head-wise dot product between the block's keys and the mean query:
\begin{equation}
  \label{eq:relevance_score}
  S(B; \ell) = \frac{1}{|B|} \sum_{u \in B} \underbrace{ \left( \frac{1}{H\sqrt{d}} \sum_{h=1}^H \bar{\mathbf{q}}_{h}^{(\ell)\top} \mathbf{k}_{u,h}^{(\ell)} \right) }_{\text{Head-averaged similarity}} .
\end{equation}

\subsection{Model-Aligned Dual Probing}
Leveraging the layer specialization observed in \cref{sec:step_layer_specialization}, we apply this scoring function at two distinct depths to curate history:

\paragraph{Text Selection (Grounding).}
At an early layer $\ell_{\text{grd}}$ (default $\ell_{\text{grd}}{=}1$), we select the top-$K_{\text{grd}}$ turns based on the relevance of their text blocks: $\mathcal{K}_{\text{grd}} = \mathrm{TopK}\big( \{ S(T_i; \ell_{\text{grd}}) \}_{i=1}^m, K_{\text{grd}} \big)$.

\paragraph{Image Selection (Synthesis).}
At a late layer $\ell_{\text{syn}}$ (default $\ell_{\text{syn}}{=}15$), we select the top-$K_{\text{img}}$ turns based on their VAE blocks to preserve visual consistency: $\mathcal{K}_{\text{syn}} = \mathrm{TopK}\big( \{ S(V_i; \ell_{\text{syn}}) \}_{i=1}^m, K_{\text{img}} \big)$.

\vspace{2pt}
We find that a small budget of $K_{\text{img}}\approx 4$ offers the optimal trade-off between providing sufficient visual reference and minimizing attention pollution.

\subsection{KV Eviction During Generation}
\label{sec:drop_uniform}
Once the sets of relevant turns $\mathcal{K}_{\text{grd}}$ and $\mathcal{K}_{\text{syn}}$ are identified via the one-shot probe, UniLongGen enforces a fixed, layer-split visibility policy throughout the subsequent generation steps.
We strictly \emph{evict} non-selected tokens from the KV cache rather than compressing them, ensuring that spurious visual outliers are completely removed from the softmax competition.
Let $\mathcal{H}^{(\ell)}$ denote the visible history at layer $\ell$. The policy is:
\begin{equation}
  \mathcal{H}^{(\ell)} =
  \begin{cases}
    \mathcal{T}_{\text{text}}(\{1\}\cup\mathcal{K}_{\text{grd}}), & \ell < \ell_{\text{syn}}, \\[2pt]
    \mathcal{T}_{\text{img}}(\{1\}\cup\mathcal{K}_{\text{syn}}), & \ell \ge \ell_{\text{syn}},
  \end{cases}
\end{equation}
where $\mathcal{T}_{\text{text}}(\cdot)$ keeps the token indices of the selected \emph{text blocks}, and $\mathcal{T}_{\text{img}}(\cdot)$ keeps the token indices of the selected \emph{VAE image blocks}.
The selected turns are always turn~$1$ plus the Top-$K$ turns ($\mathcal{K}_{\text{grd}}$ or $\mathcal{K}_{\text{syn}}$).
This split aligns with the model's functional hierarchy: early layers ground the generation in relevant text descriptions, while late layers attend to relevant visual references for synthesis.

\begin{table*}[t]
  \centering
  \small
  \setlength{\tabcolsep}{3.8pt}
  \renewcommand{\arraystretch}{1.10}

  \caption{\textbf{Systematic ablations for long interleaved text--image generation.}
    Columns are grouped into \textbf{Selection design}, \textbf{Downsampling}, and \textbf{Metrics}.
    HPS v2 / PickScore / Qual.\ measure visual quality; DINOv2 / ID Cons.\ / Style Cons.\ measure consistency. Qual., ID Cons., and Style Cons.\ are GPT-4o based evaluations.
    \emph{Within each block, we vary only one axis while keeping the rest fixed for controlled comparison.}
  Our best configuration uses one-shot dual probing (Text at Layer 1, VAE at a late layer) with a fixed, layer-split KV visibility to retain a small number of reference images ($K\approx 4$--$6$) with direct dropping of discarded tokens (no compression).}
  \label{tab:big_ablation}

  \resizebox{\textwidth}{!}{%
    \begin{tabular}{@{}L{4.55cm}
        C{1.70cm} C{1.10cm} C{1.50cm}
        C{1.30cm} C{0.50cm} C{1.20cm}
        C{1.25cm} C{1.25cm} C{0.90cm}
      C{0.90cm} C{0.90cm} C{1.30cm}@{}}

      \toprule
      \multirow{2}{*}{\textbf{Variant}} &
      \multicolumn{3}{c}{\textbf{\mbox{Selection design}}} &
      \multicolumn{3}{c}{\textbf{\mbox{Downsampling}}} &
      \multicolumn{3}{c}{\textbf{\mbox{Quality}$\uparrow$}} &
      \multicolumn{3}{c}{\textbf{\mbox{Consistency}$\uparrow$}} \\
      \cmidrule(lr){2-4}\cmidrule(lr){5-7}\cmidrule(lr){8-10}\cmidrule(lr){11-13}
      & \textbf{Signal} & \textbf{Unit} & \textbf{Budget} &
      \textbf{Discard} & \textbf{Rate} & \textbf{Interp} &
      \textbf{HPS v3} & \textbf{PickScore} & \textbf{Qual.} &
      \textbf{ID} & \textbf{Style} & \textbf{DINOv2} \\
      \midrule

      \rowcolor{blockGray}
      \multicolumn{13}{l}{\textbf{A. Baselines} \textcolor{black!55}{(no sparsification / naive heuristics)}} \\
      \midrule
      Base: Dense KV        & \NA & \NA   & \NA   & \NA & \NA & \NA & 3.1677 & 0.1943 & 5.83 & 5.49 & 3.99 & 0.3164 \\
      Sliding window (first + last $N$ imgs) & \NA & Image & $N{=}4$ & Drop & \NA & \NA & 4.4886 & 0.1969 & 6.47 & 5.73 & 4.56 & 0.3337 \\
      Sliding window (first + last $N$ imgs) & \NA & Image & $N{=}8$ & Drop & \NA & \NA & 3.9126 & 0.1957 & 6.31 & 5.41 & 4.50 & 0.3267 \\
      Sliding window (first + last $N$ imgs) & \NA & Image & $N{=}12$ & Drop & \NA & \NA & 4.0844 & 0.1960 & 6.34 & 5.82 & 4.76 & 0.3235 \\
      \addlinespace[3pt]

      \rowcolor{blockGray}
      \multicolumn{13}{l}{\textbf{B. Probe Design} \textcolor{black!55}{(fix Unit=Image, Budget=K=4, Discard=Drop)}} \\
      \midrule
      B1: Text & Text & Image & $K{=}4$ & Drop & \NA & \NA & 7.5401 & 0.2031 & 7.56 & 7.01 & 5.64 & 0.4075 \\
      B2: ViT  & ViT & Image & $K{=}4$ & Drop & \NA & \NA & 5.5275 & 0.1989 & 6.90 & 5.96 & 4.85 & 0.3451 \\
      B3: VAE  & VAE & Image & $K{=}4$ & Drop & \NA & \NA & 3.1640 & 0.1940 & 6.00 & 4.48 & 3.61 & 0.2718 \\
      \textbf{B4: layer-split: Text$\to$VAE} & Hybrid & Image & $K{=}4$ & Drop & \NA & \NA & 7.5701 & 0.2025 & 7.58 & 7.09 & 6.13 & 0.4272 \\
      \addlinespace[3pt]

      \rowcolor{blockGray}
      \multicolumn{13}{l}{\textbf{C. What to Keep (Unit \& Budget)} \textcolor{black!55}{(fix Signal = best from B; here: Hybrid)}} \\
      \midrule
      C1: Image-level retention     & Hybrid & Image & $K{=}4$  & Drop & \NA & \NA & 7.5701 & 0.2025 & 7.58 & 7.09 & 6.13 & 0.4272 \\
      C2: Image-level retention     & Hybrid & Image & $K{=}8$  & Drop & \NA & \NA & 7.3491 & 0.2022 & 7.45 & 6.95 & 5.97 & 0.4204 \\
      C3: Image-level retention     & Hybrid & Image & $K{=}12$ & Drop & \NA & \NA & 6.9994 & 0.2013 & 7.36 & 6.72 & 6.04 & 0.3853 \\
      \addlinespace[2pt]
      C4: Token-level retention  & Hybrid & Token & $K{=}4$ & Drop & \NA & \NA & 4.6454 & 0.1974 & 6.67 & 5.68 & 4.45 & 0.3598 \\
      C5: Token-level retention  & Hybrid & Token & $K{=}8$ & Drop & \NA & \NA & 4.3463 & 0.1967 & 6.65 & 5.70 & 4.52 & 0.3431 \\
      C6: Token-level retention  & Hybrid & Token & $K{=}12$ & Drop & \NA & \NA & 4.4805 & 0.1967 & 6.61 & 5.86 & 4.67 & 0.3537 \\
      \addlinespace[2pt]
      C7: Grouped token retention  & Hybrid & Tok$\times$8  & $K{=}4$ & Drop & \NA & \NA & 4.5345 & 0.1970 & 6.60 & 5.79 & 4.25 & 0.3619 \\
      C8: Grouped token retention  & Hybrid & Tok$\times$32 & $K{=}4$ & Drop & \NA & \NA & 4.6553 & 0.1971 & 6.63 & 5.69 & 4.38 & 0.3558 \\
      C9: Grouped token retention  & Hybrid & Tok$\times$128 & $K{=}4$ & Drop & \NA & \NA & 4.6371 & 0.1974 & 6.63 & 5.67 & 4.33 & 0.3512 \\
      \addlinespace[3pt]

      \rowcolor{blockGray}
      \multicolumn{13}{l}{\textbf{D. Discard Handling}  \textcolor{black!55}{(fix Signal, Unit=Image, Budget=K=4; vary compression design)}} \\
      \midrule
      D1: Drop discarded tokens                  & Hybrid & Image & $K{=}4$ & Drop       & \NA       & \NA     & 7.5701 & 0.2025 & 7.58 & 7.09 & 6.13 & 0.4272 \\
      D2: Downsample (AvgPool)  & Hybrid & Image & $K{=}4$ & Compress & $\times4$ & AvgPool & 6.9601 & 0.2018 & 7.36 & 6.92 & 5.32 & 0.3997 \\
      D3: Downsample (MaxPool)  & Hybrid & Image & $K{=}4$ & Compress & $\times4$ & MaxPool & -0.527 & 0.1895 & 4.71 & 3.85 & 3.24 & 0.2078 \\
      D5: Downsample (interp)  & Hybrid & Image & $K{=}4$ & Compress & $\times4$ & LERP    & 7.1506 & 0.2019 & 7.47 & 7.01 & 5.79 & 0.3973 \\
      D6: Downsample (interp)  & Hybrid & Image & $K{=}4$ & Compress & $\times8$ & LERP    & 7.3643 & 0.2022 & 7.54 & 6.94 & 5.55 & 0.4124 \\
      D7: Downsample (interp)  & Hybrid & Image & $K{=}4$ & Compress & $\times16$ & LERP    & 7.4855 & 0.2025 & 7.57 & 7.03 & 5.88 & 0.4118 \\
      \addlinespace[3pt]

      \rowcolor{blockGray}
      \multicolumn{13}{l}{\textbf{E. Final Configuration} 
      } 
      \\
      \midrule
      \rowcolor{oursRow}
      \textbf{$\star$ Ours (UniLongGen)} &
      \textbf{Text@L1 + VAE@L$\ell_{\text{syn}}$} &
      \textbf{Image} &
      \textbf{$K{=}4$} &
      \textbf{Drop} &
      \textbf{\NA} &
      \textbf{\NA} &
      \best{7.5701} & \best{0.2025} & \best{7.58} &
      \best{7.09} & \best{6.13} & \best{0.4272} \\
      \bottomrule
    \end{tabular}
  } %

  \vskip -0.1in
\end{table*}

\section{Experiments}
\label{sec:experiments}

\subsection{Experimental setup}
\label{sec:exp_setup}
We evaluate long-horizon interleaved image–text generation using a controlled benchmark of 50 story templates.
Each template defines a fixed narrative and a sequence of image-generation slots.
Text segments are provided verbatim, and the model is responsible only for image generation, isolating image-generation behavior under long multimodal context.
Stories contain recurring subjects appearing at multiple positions with varying scenes and viewpoints, enabling evaluation of long-range consistency.
We evaluate performance along two axes: image quality and cross-image consistency.
Image quality is measured using HPS v3, PickScore, and GPT-5.2.
Subject consistency is assessed using DINOv2 and GPT-5.2, and style consistency is evaluated using GPT-5.2.
Detailed metric definitions and evaluation protocols are provided in the appendix.

\paragraph{Additional results in the appendix.}
We defer several complementary experiments and analyses to the appendix for readability.
Generalization to a pure autoregressive unified model (Lumina-mGPT) is reported in \cref{sec:appendix_generalization}.
Further analyses connecting our method to long-context techniques and attention dynamics are in \cref{sec:appendix_analysis}.
Additional ablations and qualitative comparisons are in \cref{sec:appendix_ablations}.

\subsection{Main results and design-space ablations}
\label{sec:exp_table1}
Table~\ref{tab:big_ablation} summarizes our main results and systematic ablations for 40-image long-horizon interleaved generation.
For clarity, we structure the discussion to mirror the table blocks: \textbf{(A) baselines}, \textbf{(B) probe  design}, \textbf{(C) what to keep (unit \& budget)}, \textbf{(D) drop vs.\ compression}, and \textbf{(E) final configuration}.

\begin{figure*}[!t]
  \begin{center}
    \centerline{\includegraphics[width=0.92\linewidth]{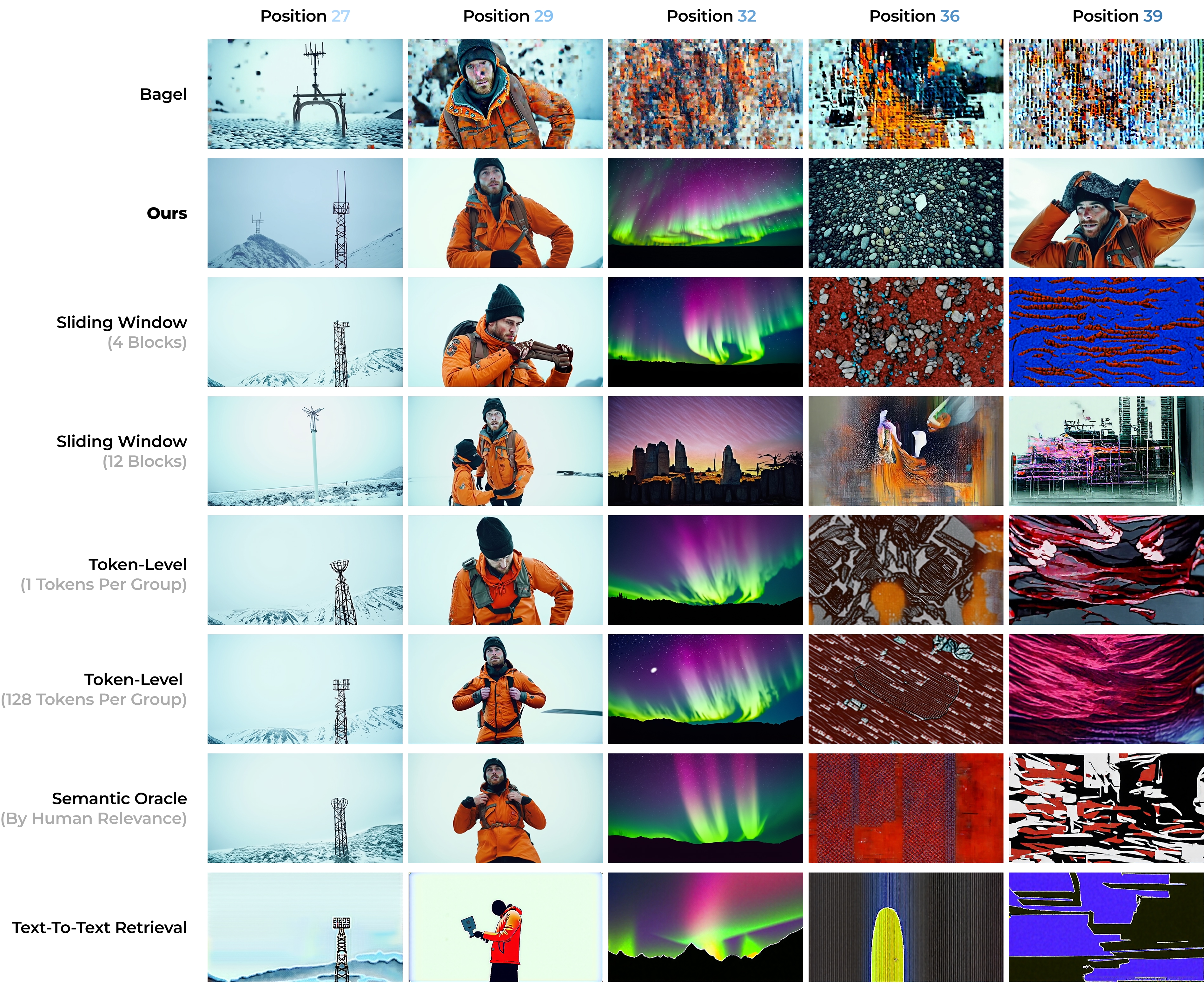}}
    \vskip -0.05in
    \caption{
      \textbf{Qualitative comparison of long-horizon stability across context-management baselines.}
      We show snapshots from the same 40-image interleaved generation at late positions (columns: 27, 29, 32, 36, 39) at $768{\times}432$ resolution.
      Rows compare the base model with dense KV (Bagel), UniLongGen, anchored sliding-window baselines (keep first anchor plus last $N$ image blocks), token-level selection, a human semantic oracle, and text-to-text retrieval.
      Dense KV and several heuristics degrade into artifacts or drift at later positions, while UniLongGen better preserves coherent synthesis in this representative example.
    }
    \label{fig:comparison}
  \end{center}
  \vskip -0.1in
\end{figure*}

\paragraph{(A) Baselines: why dense KV fails, and why recency windows are brittle.}
Dense KV collapses in the long-horizon regime (HPS v3 $=3.17$, DINOv2 $=0.316$), consistent with our diagnosis that more history can be actively harmful: attention entropy grows (\cref{fig:attn_entropy}) and key-reference mass is eroded by accumulated distractors (\cref{fig:visual_memory}).
A naive recency window partially helps by reducing the number of image events visible to attention, but remains brittle: it is misaligned with long-range story constraints and tends to drop early identity/style anchors or non-recent but high-value references.

\paragraph{(B) Probe design: one probe is not enough; depth matters.}
Fixing Unit=Image and $K{=}4$, the choice of relevance signal dominates performance.
Text @ Layer1 yields a large jump (B1: HPS v3 $=7.54$), indicating that stabilizing \emph{grounding} reduces under-conditioning and prevents drift.
However, signals must be matched to depth specialization.
ViT @ Layer1 is weaker (B2), and VAE @ Layer1 fails (B3), consistent with shallow-layer VAE interactions being dominated by high-variance patch statistics (tail-risk outliers; \cref{sec:tail_risk}).
Our \textbf{Text$\to$VAE layer-split} (B4) best balances quality and long-horizon consistency (HPS v3 $=7.57$, ID $=7.09$, Style $=6.13$): it explicitly follows the grounding-to-synthesis transition across layers (\cref{sec:step_layer_specialization}) and the zero-sum Text--VAE competition across depth (\cref{fig:token_type_ratio,fig:hist_current_competition}).

\paragraph{(C) What to keep: event-level selection matches the bottleneck and removes tail risk.}
Image-level retention (C1--C3) consistently outperforms token/grouped-token retention (C4--C9).
This directly supports the Event Bottleneck (\cref{sec:event_bottleneck}): stability is governed by how many image \emph{events} (dense key pools) remain in competition, not by preserving an arbitrary subset of tokens.
Token-level pruning leaves behind fragments of dense visual keys (and structural routing tokens), which can still produce outliers and destabilize softmax allocation.
We provide a qualitative example comparing token-level versus image-level selection under matched KV budgets in \cref{fig:token_budget}.
We also observe a small-budget optimum: increasing $K$ beyond 4 reintroduces competitors and slightly degrades quality/consistency, aligning with the ``competition'' view of failure.

\paragraph{(D) Discard handling: eviction beats summarization.}
Given the same selected references, directly dropping the rest is most reliable.
Compression (pooling/interpolation) can preserve spurious competitors and introduces distribution shift; the catastrophic max-pooling result is consistent with amplifying high-frequency artifacts that then hijack attention (\cref{sec:tail_risk}).
Overall, these results support our core mechanism: for unified generation, \emph{removing} competing visual events is safer than \emph{summarizing} them.

\paragraph{(E) Choosing the VAE probing layer $\ell_{\text{syn}}$.}
UniLongGen relies on a late-layer VAE probe to select historical image turns for synthesis (\cref{sec:method}).
Our diagnosis suggests that this probe should be taken from the generation-dominant regime (\cref{sec:step_layer_specialization}): too early and the relevance signal is noisy (tail-risk patch matches), too late and the model may overfit to step-specific details.
We vary $\ell_{\text{syn}}\in\{10,15,20\}$ while fixing the selection policy (Hybrid Text$\to$VAE), budget ($K{=}4$), and discard rule (Drop).
\Cref{tab:vae_layer_ablation} shows that performance is robust across these late layers, with a clear optimum around $\ell_{\text{syn}}{=}15$,
supporting our interpretation that long-horizon stability benefits from a synthesis-aligned relevance signal drawn from late layers, without requiring per-layer tuning.

\paragraph{Takeaway.}
Across the systematic ablations, three principles consistently emerge: \textbf{(i)} select history using model-internal, depth-aligned signals; \textbf{(ii)} prune at the event/image level with a small reference budget; \textbf{(iii)} evict discarded tokens instead of compressing them.
UniLongGen instantiates these principles via one-shot dual probing (Text@L1 and VAE@L$\ell_{\text{syn}}$) and a fixed Text$\to$VAE layer split, yielding the best overall quality--consistency trade-off.

\begin{table}[t]
  \centering
  \small
  \setlength{\tabcolsep}{4.5pt}
  \renewcommand{\arraystretch}{1.12}
  \caption{\textbf{Ablation on VAE probing layer $\ell_{\text{syn}}$.}
  We vary the layer at which the VAE attention score is computed ($\ell_{\text{syn}} \in \{10, 15, 20\}$) with Layer 15 yields the best trade-off across all metrics.}
  \label{tab:vae_layer_ablation}
  \resizebox{\columnwidth}{!}{%
    \begin{tabular}{@{}l ccc cc@{}}
      \toprule
      \textbf{Variant} &
      \textbf{HPS v3}$\uparrow$ & \textbf{PickScore}$\uparrow$ & \textbf{Qual.}$\uparrow$ &
      \textbf{Style Cons.}$\uparrow$ & \textbf{DINOv2}$\uparrow$ \\
      \midrule
      \rowcolor{blockGray}
      \multicolumn{6}{l}{\textit{Baselines}} \\
      Dense KV (Bagel) & 3.1677 & 0.1943 & 5.83 & 3.99 & 0.3164 \\
      Sliding window   & 4.4886 & 0.1969 & 6.47 & 4.56 & 0.3337 \\
      \midrule
      \rowcolor{blockGray}
      \multicolumn{6}{l}{\textit{Ours ($K{=}4$) with varying $\ell_{\text{syn}}$}} \\
      Ours, $\ell_{\text{syn}}{=}10$ & 7.4349 & 0.2019 & 7.53 & 6.12 & 0.4251 \\
      \rowcolor{oursRow}
      \textbf{Ours, $\ell_{\text{syn}}{=}15$} & \best{7.5701} & \best{0.2025} & \best{7.58} & \best{6.13} & \best{0.4272} \\
      Ours, $\ell_{\text{syn}}{=}20$ & 7.4920 & 0.2023 & 7.52 & 6.03 & 0.4239 \\
      \bottomrule
    \end{tabular}
  }
\end{table}

\subsection{Qualitative results}
\label{sec:exp_qual}
We provide qualitative comparisons on long interleaved generation (\cref{fig:comparison,fig:supp_results_1,fig:supp_results_2,fig:supp_results_3,fig:supp_results_4,fig:supp_results_6,fig:supp_results_7,fig:supp_results_5}).
\Cref{fig:comparison} highlights a representative late-horizon segment (positions 27--39) where dense-KV generation exhibits severe degradation, while UniLongGen remains visually coherent.

\begin{figure}[h]
  \begin{center}
    \centerline{\includegraphics[width=\linewidth]{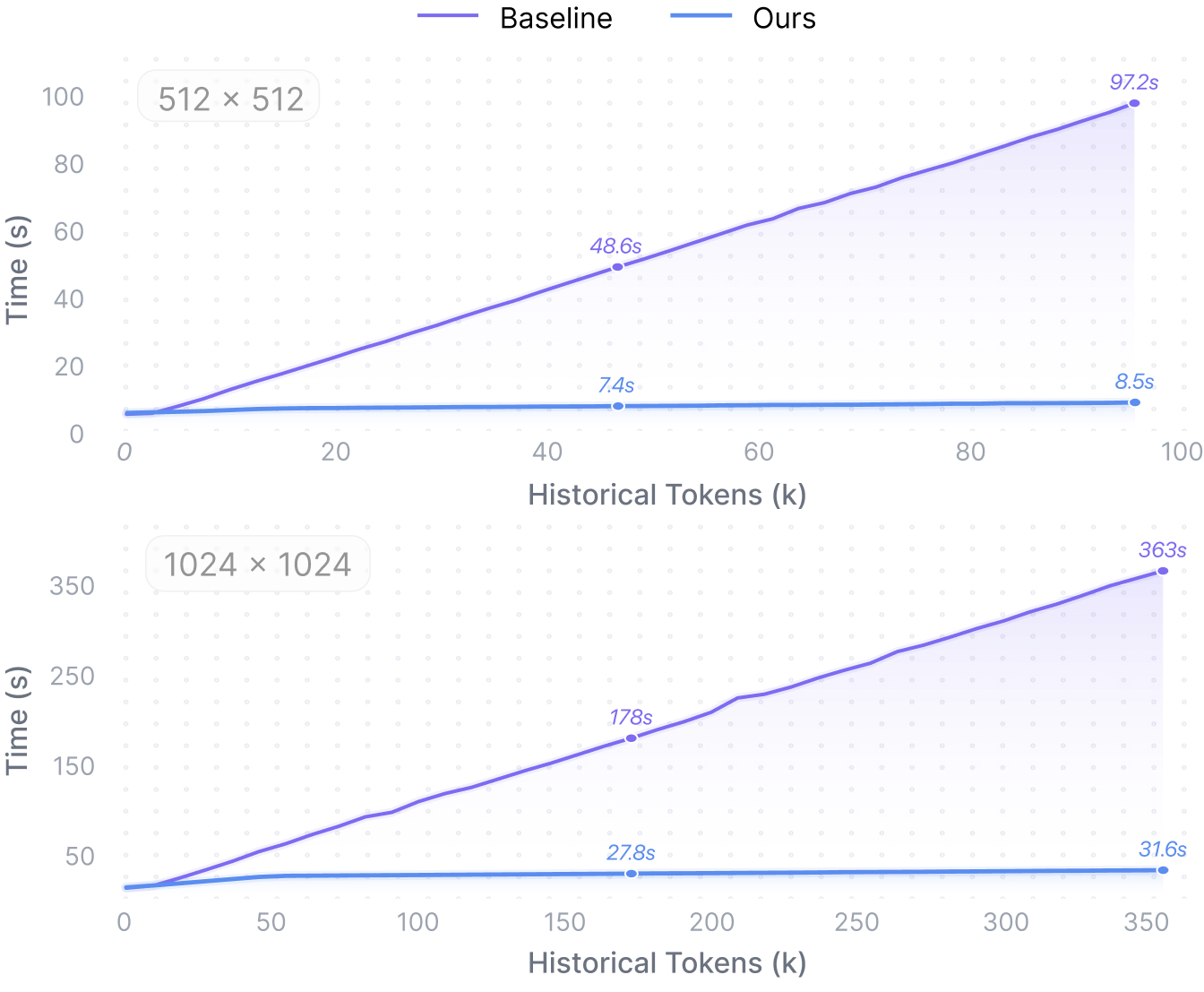}}
    \caption{
      \textbf{Runtime scaling vs.\ historical context (no CFG).}
      Total flow-matching time (seconds; summed over steps) to generate one image at two resolutions (top: $512{\times}512$, bottom: $1024{\times}1024$) as a function of historical-context tokens (k).
      Dense KV slows down with longer history, while UniLongGen remains nearly flat due to KV eviction, achieving up to $\sim$11$\times$ speedup at long contexts.
    }
    \label{fig:run_time}
  \end{center}
  \vskip -0.3in
\end{figure}

\subsection{Efficiency Comparison}
\label{sec:appendix_efficiency}
We report the average visible KV length (KV\%) and end-to-end wall-clock time per image (summed over all flow-matching steps), with classifier-free guidance disabled (\cref{fig:run_time}).
Dense KV time grows roughly linearly with historical tokens, since each step attends over the full history.
UniLongGen instead evicts non-selected history and keeps only a small, fixed set of reference turns visible, making runtime largely insensitive to raw history length: at $512{\times}512$ with 100k tokens, Dense KV takes 97.2s vs.\ 8.5s; at $1024{\times}1024$ with 350k tokens, 363s vs.\ 31.6s—up to $\sim$11$\times$ speedup.

\subsection{Qualitative late-horizon comparison.}
\Cref{fig:comparison} provides a side-by-side qualitative comparison across representative context-management baselines at late positions in the same 40-image sequence.
It illustrates common failure modes under long histories (e.g., artifact explosion, semantic drift, or collapsing into repetitive textures) and shows that UniLongGen can maintain coherent synthesis later into the sequence in this example, consistent with the quantitative gaps reported in our ablations.

\subsection{Model-aligned beats oracle semantic retention}
\label{sec:exp_oracle_semantic}
An intuitive baseline is to keep the ``most relevant'' history by an external (human/oracle) criterion---\eg, retaining the historical images that contain the target character or attributes specified by the prompt.
Surprisingly, we find that such oracle semantic retention can underperform our attention-based selection (\cref{tab:oracle_vs_attn}).
We further test a text-only matching baseline that ranks historical turns by the similarity between the current image's text block and each historical text block, then keeps the images from the top-$K$ ranked turns; this performs poorly, reinforcing that semantic/text relevance alone is insufficient for stable long-horizon synthesis.
This highlights a key asymmetry in long-horizon \emph{generation}: what is semantically relevant to humans is not always what the model needs for stable synthesis.

\paragraph{Why oracle relevance can hurt.}
(1) Relevance mismatch. ``Relevant'' for generation is not purely semantic: the model may depend on recent context for pose/layout/lighting continuity, or on specific reference views that best match its current latent trajectory.
(2) Distribution shift from hard masking. Any pruning changes the softmax normalization and can re-route attention mass.
Forcing a human-defined subset can create an unnatural context mixture, amplifying spurious dependencies and destabilizing identity/style even if the retained images appear correct.
In contrast, attention-based selection is \emph{model-aligned}: it preserves the subset of historical image blocks the model itself prefers under its internal representations, yielding better long-horizon fidelity.

\begin{implication}[Model-Alignment Over Semantic Intuition]
  Model-aligned selection can outperform semantic oracles, highlighting a gap between \emph{semantic relevance} and \emph{generative utility}. Therefore, we derive curation from the model's internal relevance signals rather than external heuristics.
\end{implication}

\begin{table}[t]
  \centering
  \small
  \setlength{\tabcolsep}{5.5pt}
  \renewcommand{\arraystretch}{1.15}
  \caption{\textbf{Semantic oracle vs.\ attention-based selection.}
  ``Semantic oracle'' keeps the same budget ($K$) of historical images judged most relevant by human annotation; 
  ``Text-block matching'' ranks historical turns by the similarity between the current image's text block and each historical text block, then keeps the images from the top-$K$ turns.
  Our method selects images using model-internal attention signals.}
  \label{tab:oracle_vs_attn}
  \vspace{-3pt}
  \resizebox{\columnwidth}{!}{%
    \begin{tabular}{@{}lcccc@{}}
      \toprule
      \textbf{Variant} & \textbf{HPS v3}$\uparrow$ & \textbf{Qual.}$\uparrow$ & \textbf{Style}$\uparrow$ & \textbf{DINOv2}$\uparrow$ \\
      \midrule
      Dense KV (no pruning) & 3.1677 & 5.83 & 3.99 & 0.3164 \\
      Semantic oracle (top-$K$ by human relevance) & 5.1158 & 6.77 & 5.1 & 0.4205 \\
      Text-block matching (current $\rightarrow$ history, top-$K$) & 1.7063 & 5.35 & 3.06 & 0.3094 \\
      \rowcolor{oursRow}
      \textbf{UniLongGen (attention-based top-$K$)} & \best{7.5701} & \best{7.58} & \best{6.13} & \best{0.4272} \\
      \bottomrule
    \end{tabular}
  }%
  \vskip -0.05in
\end{table}

\section{Conclusion}
\label{sec:conclusion}
Unified multimodal models promise long-form interleaved generation, yet current systems hit a reliability cliff as image events accumulate.
We show this is not a generic long-context limitation: dense visual history creates event-level competition and heavy-tailed outliers that pollute attention and corrupt subsequent synthesis.
We contribute (i) systematic evidence for an \emph{Event Bottleneck}, where breakdown tracks image events more than tokens; (ii) a mechanistic diagnosis of tail-risk hijacking and depth-wise specialization; and (iii) \textbf{UniLongGen}, a training-free, model-aligned context curation policy that uses one-shot probing and a Text$\to$VAE layer split to evict competing history.
Across long-horizon benchmarks, UniLongGen improves quality and identity/style consistency while reducing KV footprint.

\begin{figure*}[tp]
  \begin{center}
    \centerline{\includegraphics[width=\linewidth]{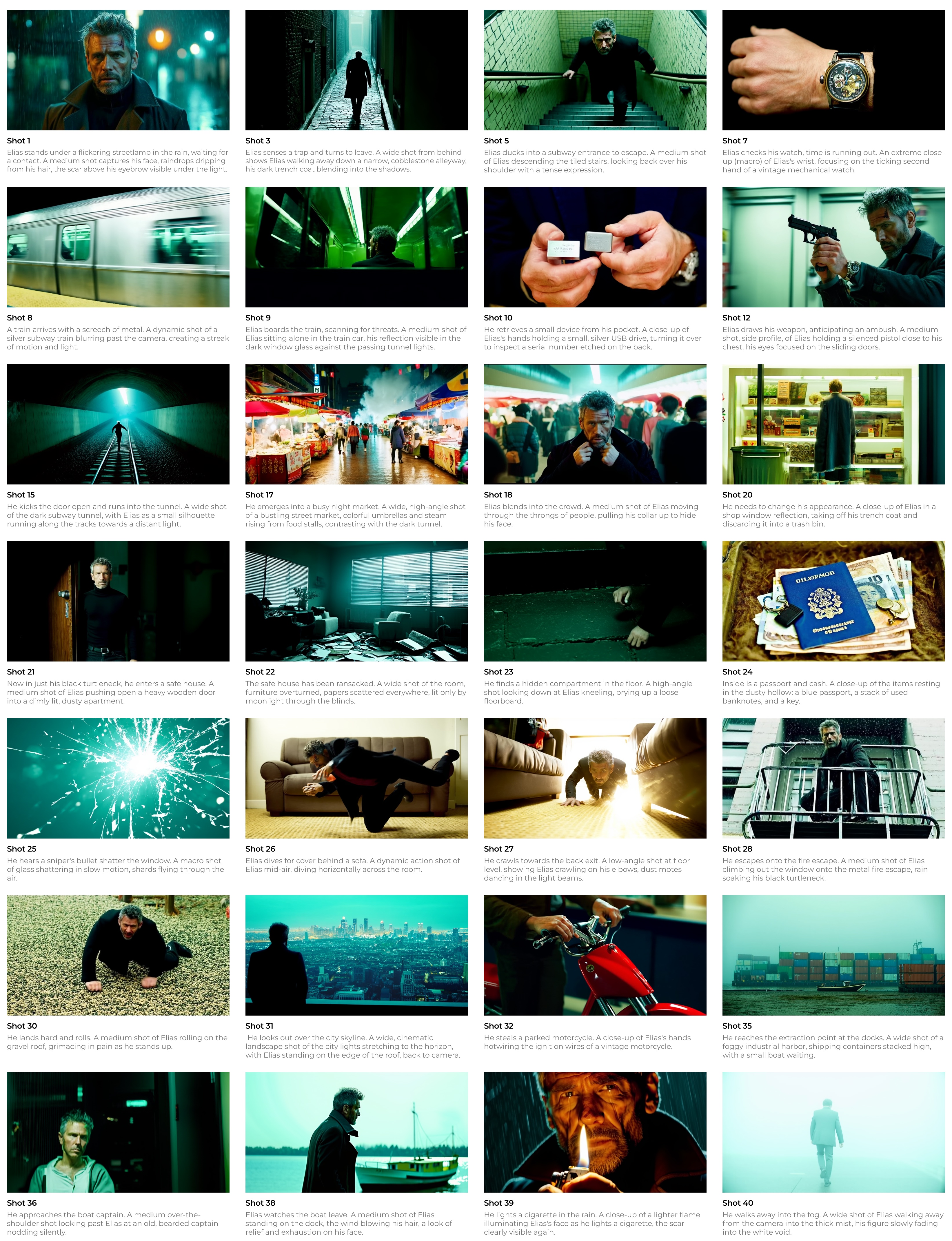}}
    \caption{Additional qualitative results on long-horizon interleaved generation (1/7).}
    \label{fig:supp_results_1}
  \end{center}
  \vskip -0.2in
\end{figure*}

\begin{figure*}[tp]
  \begin{center}
    \centerline{\includegraphics[width=\linewidth]{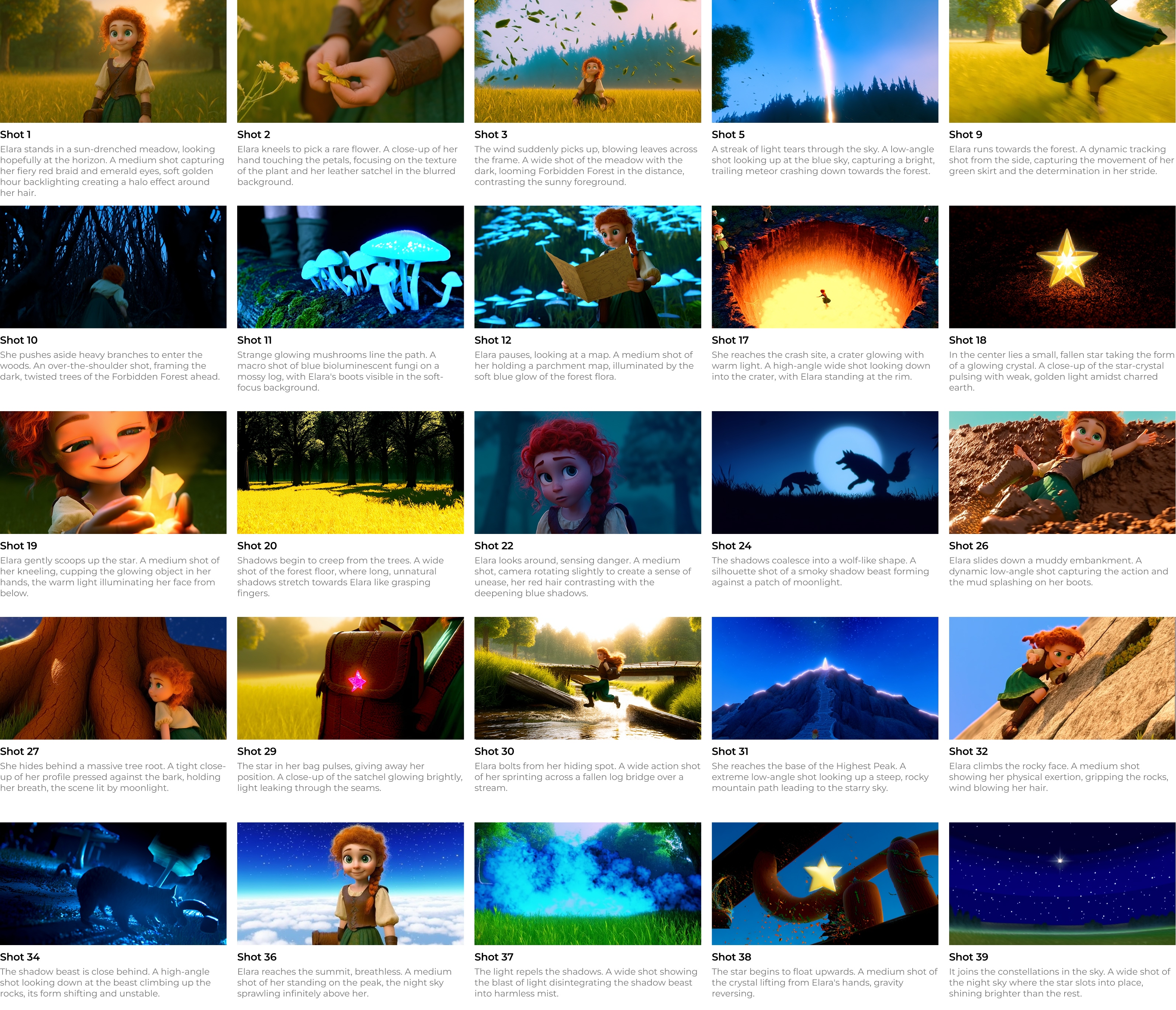}}
    \caption{Additional qualitative results on long-horizon interleaved generation (2/7).}
    \label{fig:supp_results_2}
  \end{center}
  \vskip -0.2in
\end{figure*}

\begin{figure*}[tp]
  \begin{center}
    \centerline{\includegraphics[width=\linewidth]{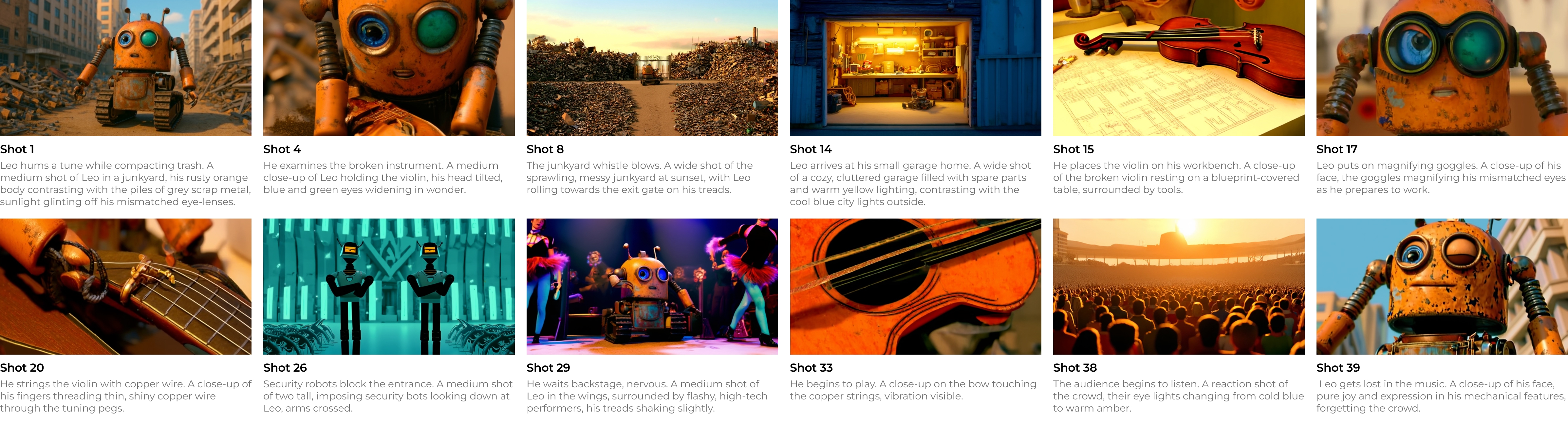}}
    \caption{Additional qualitative results on long-horizon interleaved generation (3/7).}
    \label{fig:supp_results_3}
  \end{center}
  \vskip -0.2in
\end{figure*}

\begin{figure*}[tp]
  \begin{center}
    \centerline{\includegraphics[width=\linewidth]{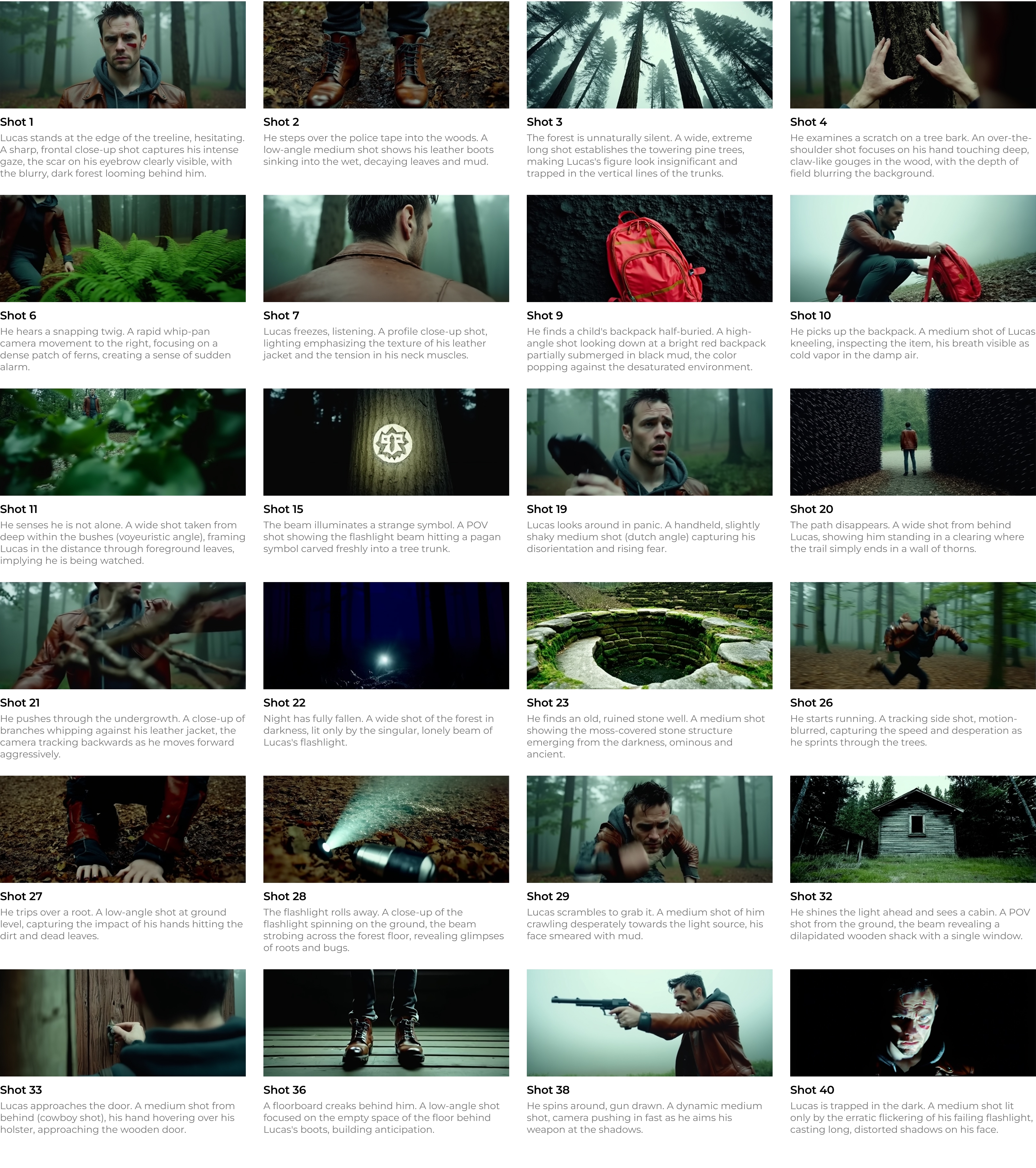}}
    \caption{Additional qualitative results on long-horizon interleaved generation (4/7).}
    \label{fig:supp_results_4}
  \end{center}
  \vskip -0.2in
\end{figure*}

\begin{figure*}[tp]
  \begin{center}
    \centerline{\includegraphics[width=0.93\linewidth]{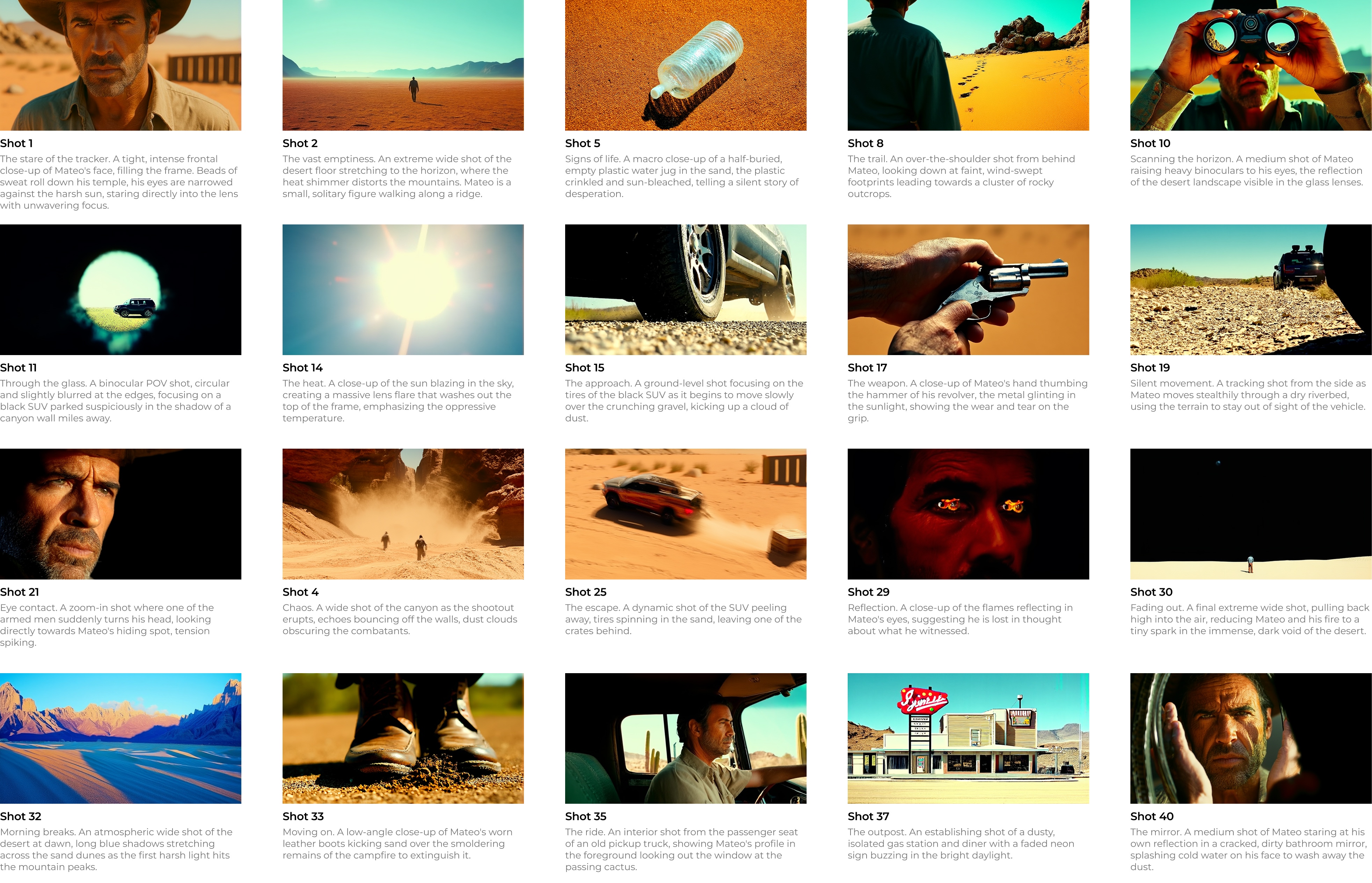}}
    \caption{Additional qualitative results on long-horizon interleaved generation (5/7).}
    \label{fig:supp_results_6}
  \end{center}
  \vskip -0.2in
\end{figure*}

\begin{figure*}[tp]
  \begin{center}
    \centerline{\includegraphics[width=0.93\linewidth]{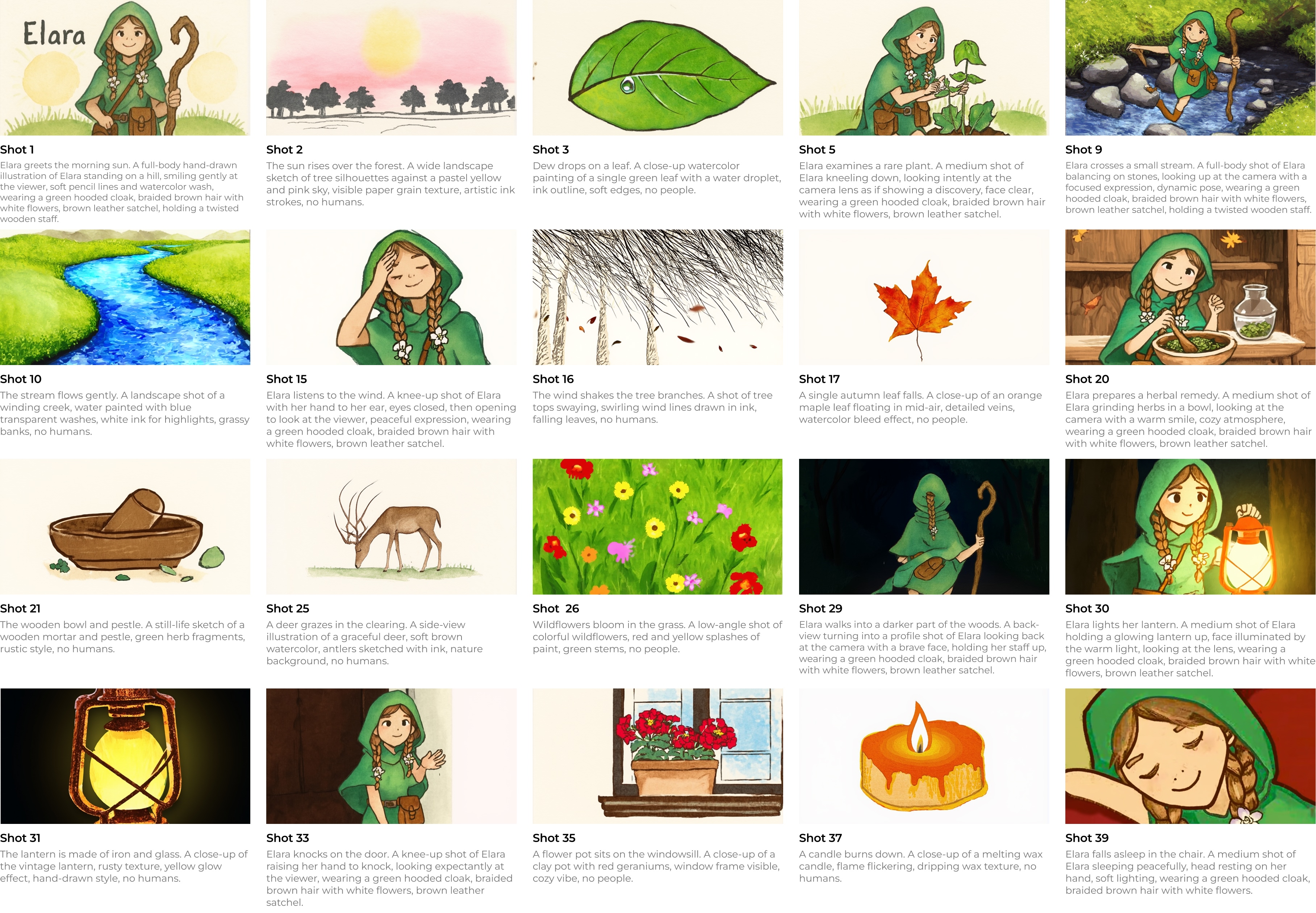}}
    \caption{Additional qualitative results on long-horizon interleaved generation (6/7).}
    \label{fig:supp_results_7}
  \end{center}
  \vskip -0.2in
\end{figure*}

\begin{figure*}[tp]
  \begin{center}
    \centerline{\includegraphics[width=\linewidth]{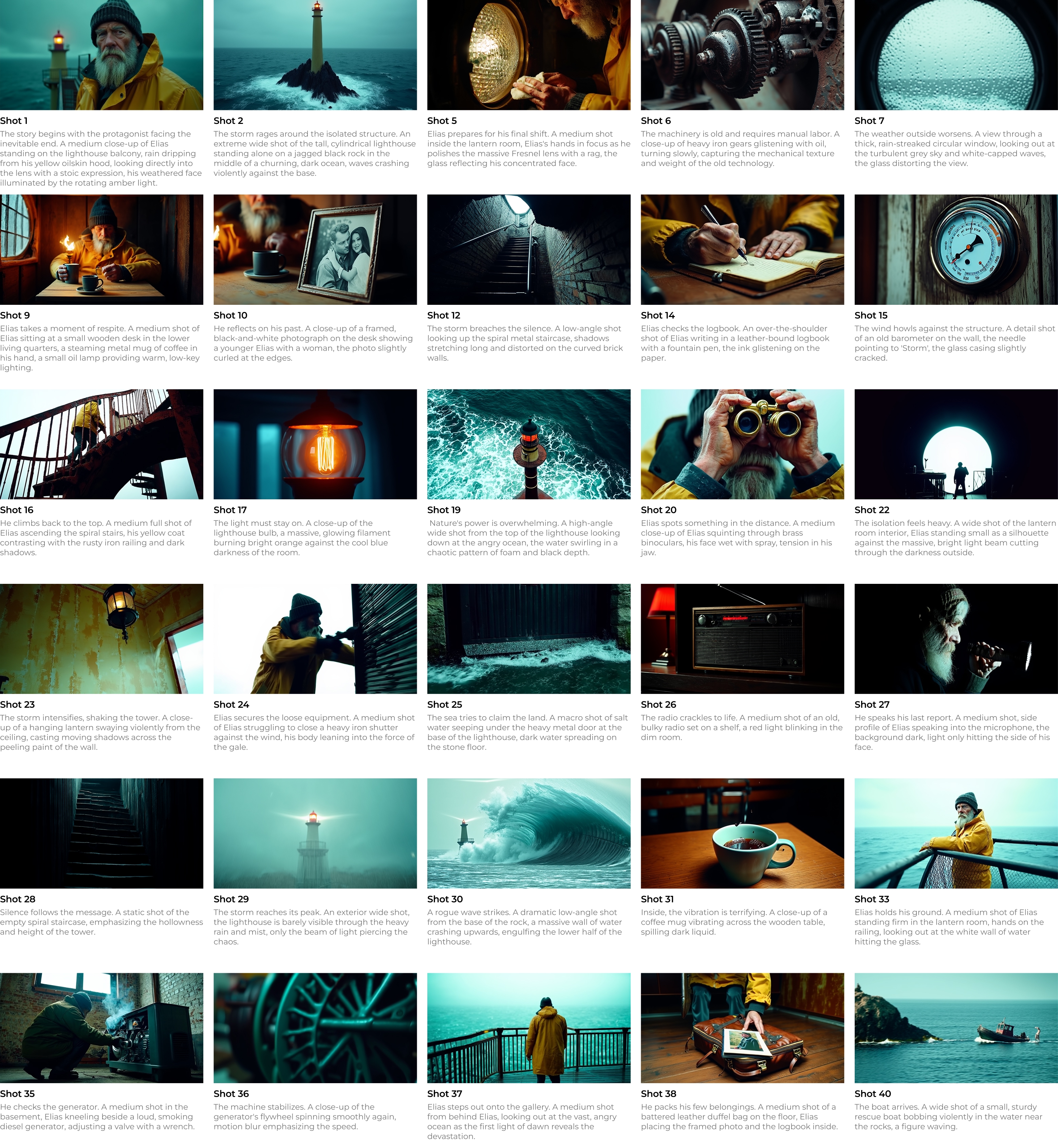}}
    \caption{Additional qualitative results on long-horizon interleaved generation (7/7).}
    \label{fig:supp_results_5}
  \end{center}
  \vskip -0.2in
\end{figure*}

\clearpage

\bibliography{example_paper}

\begin{thebibliography}{32}
\providecommand{\natexlab}[1]{#1}
\providecommand{\url}[1]{\texttt{#1}}
\expandafter\ifx\csname urlstyle\endcsname\relax
  \providecommand{\doi}[1]{doi: #1}\else
  \providecommand{\doi}{doi: \begingroup \urlstyle{rm}\Url}\fi

\bibitem[Cao et~al.(2025)Cao, Chen, Chen, Cheng, Cui, Deng, Dong, Gong, Gu, Gu,
  et~al.]{hunyuanimage3_2025}
Cao, S., Chen, H., Chen, P., Cheng, Y., Cui, Y., Deng, X., Dong, Y., Gong, K.,
  Gu, T., Gu, X., et~al.
\newblock Hunyuanimage 3.0 technical report.
\newblock \emph{arXiv preprint arXiv:2509.23951}, 2025.

\bibitem[Chen et~al.(2025{\natexlab{a}})Chen, Xu, Pan, Hu, Qin, Goldstein,
  Huang, Zhou, Xie, Savarese, et~al.]{blip3o2025}
Chen, J., Xu, Z., Pan, X., Hu, Y., Qin, C., Goldstein, T., Huang, L., Zhou, T.,
  Xie, S., Savarese, S., et~al.
\newblock Blip3-o: A family of fully open unified multimodal
  models-architecture, training and dataset.
\newblock \emph{arXiv preprint arXiv:2505.09568}, 2025{\natexlab{a}}.

\bibitem[Chen et~al.(2026)Chen, He, Fu, Wan, Gai, and Ye]{vino2026}
Chen, J., He, T., Fu, Z., Wan, P., Gai, K., and Ye, W.
\newblock Vino: A unified visual generator with interleaved omnimodal context.
\newblock \emph{arXiv preprint arXiv:2601.02358}, 2026.

\bibitem[Chen et~al.(2025{\natexlab{b}})Chen, Li, Yang, Wen, Yang, Gao, Wu, and
  Chen]{comm2024}
Chen, W., Li, L., Yang, Y., Wen, B., Yang, F., Gao, T., Wu, Y., and Chen, L.
\newblock Comm: A coherent interleaved image-text dataset for multimodal
  understanding and generation.
\newblock In \emph{Proceedings of the Computer Vision and Pattern Recognition
  Conference}, pp.\  8073--8082, 2025{\natexlab{b}}.

\bibitem[Deng et~al.(2025)Deng, Zhu, Li, Gou, Li, Wang, Zhong, Yu, Nie, Song,
  et~al.]{bagel2025}
Deng, C., Zhu, D., Li, K., Gou, C., Li, F., Wang, Z., Zhong, S., Yu, W., Nie,
  X., Song, Z., et~al.
\newblock Emerging properties in unified multimodal pretraining.
\newblock \emph{arXiv preprint arXiv:2505.14683}, 2025.

\bibitem[Dong et~al.(2023)Dong, Han, Peng, Qi, Ge, Yang, Zhao, Sun, Zhou, Wei,
  et~al.]{dreamllm2023}
Dong, R., Han, C., Peng, Y., Qi, Z., Ge, Z., Yang, J., Zhao, L., Sun, J., Zhou,
  H., Wei, H., et~al.
\newblock Dreamllm: Synergistic multimodal comprehension and creation.
\newblock \emph{arXiv preprint arXiv:2309.11499}, 2023.

\bibitem[Ge et~al.(2024)Ge, Zhao, Zhu, Ge, Yi, Song, Li, Ding, and
  Shan]{seedx2024}
Ge, Y., Zhao, S., Zhu, J., Ge, Y., Yi, K., Song, L., Li, C., Ding, X., and
  Shan, Y.
\newblock Seed-x: Multimodal models with unified multi-granularity
  comprehension and generation.
\newblock \emph{arXiv preprint arXiv:2404.14396}, 2024.

\bibitem[Hao et~al.(2025)Hao, Liu, Xiao, Huang, and Yu]{unix2025}
Hao, J., Liu, H., Xiao, X., Huang, Q., and Yu, J.
\newblock Uni-x: Mitigating modality conflict with a two-end-separated
  architecture for unified multimodal models.
\newblock \emph{arXiv preprint arXiv:2509.24365}, 2025.

\bibitem[Jiao et~al.(2025)Jiao, Qiu, Jie, Chen, Chen, Ma, and
  Jiang]{unitoken2025}
Jiao, Y., Qiu, H., Jie, Z., Chen, S., Chen, J., Ma, L., and Jiang, Y.-G.
\newblock Unitoken: Harmonizing multimodal understanding and generation through
  unified visual encoding.
\newblock In \emph{Proceedings of the Computer Vision and Pattern Recognition
  Conference}, pp.\  3600--3610, 2025.

\bibitem[Kou et~al.(2024)Kou, Jin, Liu, Liu, Ma, Jia, Chen, Jiang, and
  Deng]{orthus2024}
Kou, S., Jin, J., Liu, Z., Liu, C., Ma, Y., Jia, J., Chen, Q., Jiang, P., and
  Deng, Z.
\newblock Orthus: Autoregressive interleaved image-text generation with
  modality-specific heads.
\newblock \emph{arXiv preprint arXiv:2412.00127}, 2024.

\bibitem[Lauren{\c{c}}on et~al.(2023)Lauren{\c{c}}on, Saulnier, Tronchon,
  Bekman, Singh, Lozhkov, Wang, Karamcheti, Rush, Kiela, et~al.]{obelics2023}
Lauren{\c{c}}on, H., Saulnier, L., Tronchon, L., Bekman, S., Singh, A.,
  Lozhkov, A., Wang, T., Karamcheti, S., Rush, A., Kiela, D., et~al.
\newblock Obelics: An open web-scale filtered dataset of interleaved image-text
  documents.
\newblock \emph{Advances in Neural Information Processing Systems},
  36:\penalty0 71683--71702, 2023.

\bibitem[Li et~al.(2025)Li, Peng, Wang, Peng, Chen, Weng, Wang, Cai, Dai, and
  Xiong]{onecat2025}
Li, H., Peng, X., Wang, Y., Peng, Z., Chen, X., Weng, R., Wang, J., Cai, X.,
  Dai, W., and Xiong, H.
\newblock Onecat: Decoder-only auto-regressive model for unified understanding
  and generation.
\newblock \emph{arXiv preprint arXiv:2509.03498}, 2025.

\bibitem[Liao et~al.(2025)Liao, Liu, Wang, Luo, Zhang, Zhao, Wu, Li, Tian, and
  Huang]{mogao2025}
Liao, C., Liu, L., Wang, X., Luo, Z., Zhang, X., Zhao, W., Wu, J., Li, L.,
  Tian, Z., and Huang, W.
\newblock Mogao: An omni foundation model for interleaved multi-modal
  generation.
\newblock \emph{arXiv preprint arXiv:2505.05472}, 2025.

\bibitem[Lin et~al.(2025)Lin, Li, Cheng, Niu, Ye, He, Yuan, Yu, Wang, Ge,
  et~al.]{uniworld2025}
Lin, B., Li, Z., Cheng, X., Niu, Y., Ye, Y., He, X., Yuan, S., Yu, W., Wang,
  S., Ge, Y., et~al.
\newblock Uniworld: High-resolution semantic encoders for unified visual
  understanding and generation.
\newblock \emph{arXiv preprint arXiv:2506.03147}, 2025.

\bibitem[Liu et~al.(2024)Liu, Zhao, Zhuo, Lin, Qiao, Li, and
  Gao]{liu2024lumina-mgpt}
Liu, D., Zhao, S., Zhuo, L., Lin, W., Qiao, Y., Li, H., and Gao, P.
\newblock Lumina-mgpt: Illuminate flexible photorealistic text-to-image
  generation with multimodal generative pretraining, 2024.
\newblock URL \url{https://arxiv.org/abs/2408.02657}.

\bibitem[Ma et~al.(2025)Ma, Liu, Chen, Liu, Wu, Wu, Pan, Xie, Zhang, Yu,
  et~al.]{janusflow2024}
Ma, Y., Liu, X., Chen, X., Liu, W., Wu, C., Wu, Z., Pan, Z., Xie, Z., Zhang,
  H., Yu, X., et~al.
\newblock Janusflow: Harmonizing autoregression and rectified flow for unified
  multimodal understanding and generation.
\newblock In \emph{Proceedings of the Computer Vision and Pattern Recognition
  Conference}, pp.\  7739--7751, 2025.

\bibitem[Pan et~al.(2023)Pan, Dong, Huang, Peng, Chen, and Wei]{kosmosg2023}
Pan, X., Dong, L., Huang, S., Peng, Z., Chen, W., and Wei, F.
\newblock Kosmos-g: Generating images in context with multimodal large language
  models.
\newblock \emph{arXiv preprint arXiv:2310.02992}, 2023.

\bibitem[Pan et~al.(2025)Pan, Shukla, Singh, Zhao, Mishra, Wang, Xu, Chen, Li,
  Juefei-Xu, et~al.]{metaquery2025}
Pan, X., Shukla, S.~N., Singh, A., Zhao, Z., Mishra, S.~K., Wang, J., Xu, Z.,
  Chen, J., Li, K., Juefei-Xu, F., et~al.
\newblock Transfer between modalities with metaqueries.
\newblock \emph{arXiv preprint arXiv:2504.06256}, 2025.

\bibitem[Qu et~al.(2025)Qu, Zhang, Liu, Wang, Jiang, Gao, Ye, Du, Yuan, and
  Wu]{tokenflow2024}
Qu, L., Zhang, H., Liu, Y., Wang, X., Jiang, Y., Gao, Y., Ye, H., Du, D.~K.,
  Yuan, Z., and Wu, X.
\newblock Tokenflow: Unified image tokenizer for multimodal understanding and
  generation.
\newblock In \emph{Proceedings of the Computer Vision and Pattern Recognition
  Conference}, pp.\  2545--2555, 2025.

\bibitem[Shi et~al.(2024)Shi, Han, Zhou, Liang, Lin, Zettlemoyer, and
  Yu]{lmfusion2024}
Shi, W., Han, X., Zhou, C., Liang, W., Lin, X.~V., Zettlemoyer, L., and Yu, L.
\newblock Lmfusion: Adapting pretrained language models for multimodal
  generation.
\newblock \emph{arXiv preprint arXiv:2412.15188}, 2024.

\bibitem[Sun et~al.(2024)Sun, Cui, Zhang, Zhang, Yu, Wang, Rao, Liu, Huang, and
  Wang]{emu2_2023}
Sun, Q., Cui, Y., Zhang, X., Zhang, F., Yu, Q., Wang, Y., Rao, Y., Liu, J.,
  Huang, T., and Wang, X.
\newblock Generative multimodal models are in-context learners.
\newblock In \emph{Proceedings of the IEEE/CVF Conference on Computer Vision
  and Pattern Recognition}, pp.\  14398--14409, 2024.

\bibitem[Team(2024)]{chameleon2024}
Team, C.
\newblock Chameleon: Mixed-modal early-fusion foundation models.
\newblock \emph{arXiv preprint arXiv:2405.09818}, 2024.

\bibitem[Wang et~al.(2025{\natexlab{a}})Wang, Zhao, Zhang, Cao, Zhan, Duan, Lu,
  Fu, Chen, Zhao, et~al.]{ovisu1_2025}
Wang, G.-H., Zhao, S., Zhang, X., Cao, L., Zhan, P., Duan, L., Lu, S., Fu, M.,
  Chen, X., Zhao, J., et~al.
\newblock Ovis-u1 technical report.
\newblock \emph{arXiv preprint arXiv:2506.23044}, 2025{\natexlab{a}}.

\bibitem[Wang et~al.(2024)Wang, Zhang, Luo, Sun, Cui, Wang, Zhang, Wang, Li,
  Yu, et~al.]{emu3_2024}
Wang, X., Zhang, X., Luo, Z., Sun, Q., Cui, Y., Wang, J., Zhang, F., Wang, Y.,
  Li, Z., Yu, Q., et~al.
\newblock Emu3: Next-token prediction is all you need.
\newblock \emph{arXiv preprint arXiv:2409.18869}, 2024.

\bibitem[Wang et~al.(2025{\natexlab{b}})Wang, Zhang, Zhang, Lin, Zhou, Liu,
  Zhang, Li, Liu, Zheng, et~al.]{hbridge2025}
Wang, X., Zhang, Z., Zhang, H., Lin, Z., Zhou, Y., Liu, Q., Zhang, S., Li, Y.,
  Liu, S., Zheng, H., et~al.
\newblock Hbridge: H-shape bridging of heterogeneous experts for unified
  multimodal understanding and generation.
\newblock \emph{arXiv preprint arXiv:2511.20520}, 2025{\natexlab{b}}.

\bibitem[Wu et~al.(2025)Wu, Zheng, Yan, Xiao, Luo, Wang, Li, Jiang, Liu, Zhou,
  et~al.]{omnigen2_2025}
Wu, C., Zheng, P., Yan, R., Xiao, S., Luo, X., Wang, Y., Li, W., Jiang, X.,
  Liu, Y., Zhou, J., et~al.
\newblock Omnigen2: Exploration to advanced multimodal generation.
\newblock \emph{arXiv preprint arXiv:2506.18871}, 2025.

\bibitem[Wu et~al.(2024)Wu, Fei, Qu, Ji, and Chua]{nextgpt2023}
Wu, S., Fei, H., Qu, L., Ji, W., and Chua, T.-S.
\newblock Next-gpt: Any-to-any multimodal llm.
\newblock In \emph{Forty-first International Conference on Machine Learning},
  2024.

\bibitem[Xie et~al.()Xie, Mao, Bai, Zhang, Wang, Lin, Gu, Chen, Yang, and
  Shou]{showo2_2025}
Xie, J., Mao, W., Bai, Z., Zhang, D.~J., Wang, W., Lin, K.~Q., Gu, Y., Chen,
  Z., Yang, Z., and Shou, M.~Z.
\newblock Show-o: One single transformer to unify multimodal understanding and
  generation.
\newblock In \emph{The Thirteenth International Conference on Learning
  Representations}.

\bibitem[Xie et~al.(2024)Xie, Mao, Bai, Zhang, Wang, Lin, Gu, Chen, Yang, and
  Shou]{showo2024}
Xie, J., Mao, W., Bai, Z., Zhang, D.~J., Wang, W., Lin, K.~Q., Gu, Y., Chen,
  Z., Yang, Z., and Shou, M.~Z.
\newblock Show-o: One single transformer to unify multimodal understanding and
  generation.
\newblock \emph{arXiv preprint arXiv:2408.12528}, 2024.

\bibitem[Zhang et~al.(2026)Zhang, Qu, Liu, Chen, Song, Dong, Sun, Li, Wang,
  Jiang, et~al.]{nextflow2026}
Zhang, H., Qu, L., Liu, Y., Chen, H., Song, Y., Dong, Y., Sun, S., Li, X.,
  Wang, X., Jiang, Y., et~al.
\newblock Nextflow: Unified sequential modeling activates multimodal
  understanding and generation.
\newblock \emph{arXiv preprint arXiv:2601.02204}, 2026.

\bibitem[Zheng et~al.(2023)Zheng, He, and Wang]{minigpt5_2023}
Zheng, K., He, X., and Wang, X.~E.
\newblock Minigpt-5: Interleaved vision-and-language generation via generative
  vokens.
\newblock \emph{arXiv preprint arXiv:2310.02239}, 2023.

\bibitem[Zhou et~al.(2024)Zhou, Yu, Babu, Tirumala, Yasunaga, Shamis, Kahn, Ma,
  Zettlemoyer, and Levy]{transfusion2024}
Zhou, C., Yu, L., Babu, A., Tirumala, K., Yasunaga, M., Shamis, L., Kahn, J.,
  Ma, X., Zettlemoyer, L., and Levy, O.
\newblock Transfusion: Predict the next token and diffuse images with one
  multi-modal model.
\newblock \emph{arXiv preprint arXiv:2408.11039}, 2024.

\end{thebibliography}
\bibliographystyle{icml2026}

\newpage
\appendix
\clearpage

\clearpage

\section{Additional Analysis}
\label{sec:appendix_analysis}

\subsection{Depth-wise modality flow and Text--VAE competition in history}
\label{sec:appendix_hist_competition}
\begin{figure}[h]
  \centering
  \includegraphics[width=\columnwidth]{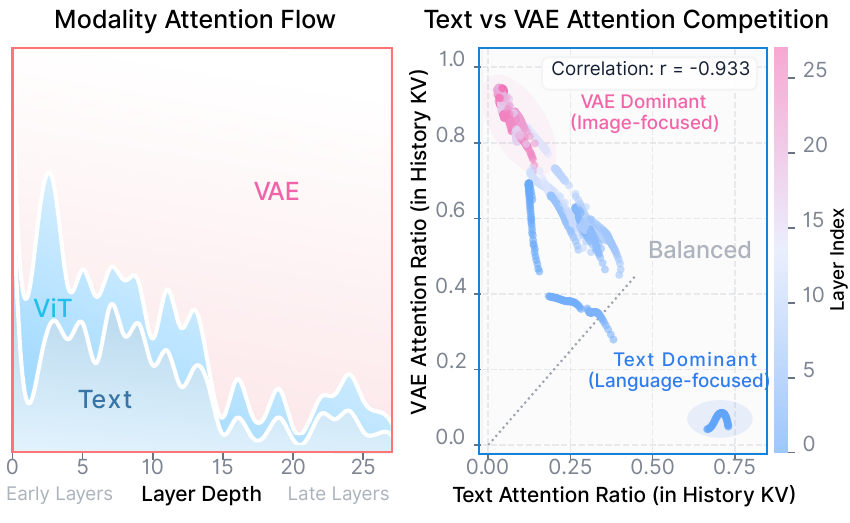}
  \caption{\textbf{Modality attention flow and Text--VAE competition across depth.}
    \emph{Left:} Modality attention ratios (Text/ViT/VAE) over Transformer depth, showing a shift from early-layer multimodal processing toward VAE-dominant late-layer image synthesis.
  \emph{Right:} Text attention ratio versus VAE attention ratio in the historical KV across layers (colored by layer index), exhibiting a strong negative correlation ($r\approx-0.93$) that indicates zero-sum competition between language grounding and image-token conditioning.}
  \label{fig:hist_current_competition}
\end{figure}

\Cref{fig:hist_current_competition} provides two complementary views of how unified generators allocate attention across modalities and depth.
The left panel shows a depth-wise shift in modality attention: early layers allocate a larger fraction of attention to text (and ViT) tokens, while later layers become increasingly VAE-dominant, consistent with a transition from multimodal grounding/routing to synthesis-centric computation.
The right panel focuses specifically on \emph{historical KV} attention and reveals a strong negative correlation between the text-attention ratio and the VAE-attention ratio across layers (colored by layer index).
This pattern suggests a near zero-sum competition within the softmax budget: allocating more attention to language grounding necessarily reduces the budget available for image-token conditioning, and vice versa.
In the long-horizon regime, where history can contain dense VAE blocks from many prior images, this competition motivates depth-aware policies (e.g., a coarse Text$\to$VAE layer split) that align visibility with the layer's functional role, rather than forcing a single uniform conditioning mixture across all depths.

\subsection{Step-wise reallocation: history vs.\ current attention}
\label{sec:appendix_stepwise_attn}
\begin{figure}[h]
  \centering
  \includegraphics[width=1\columnwidth]{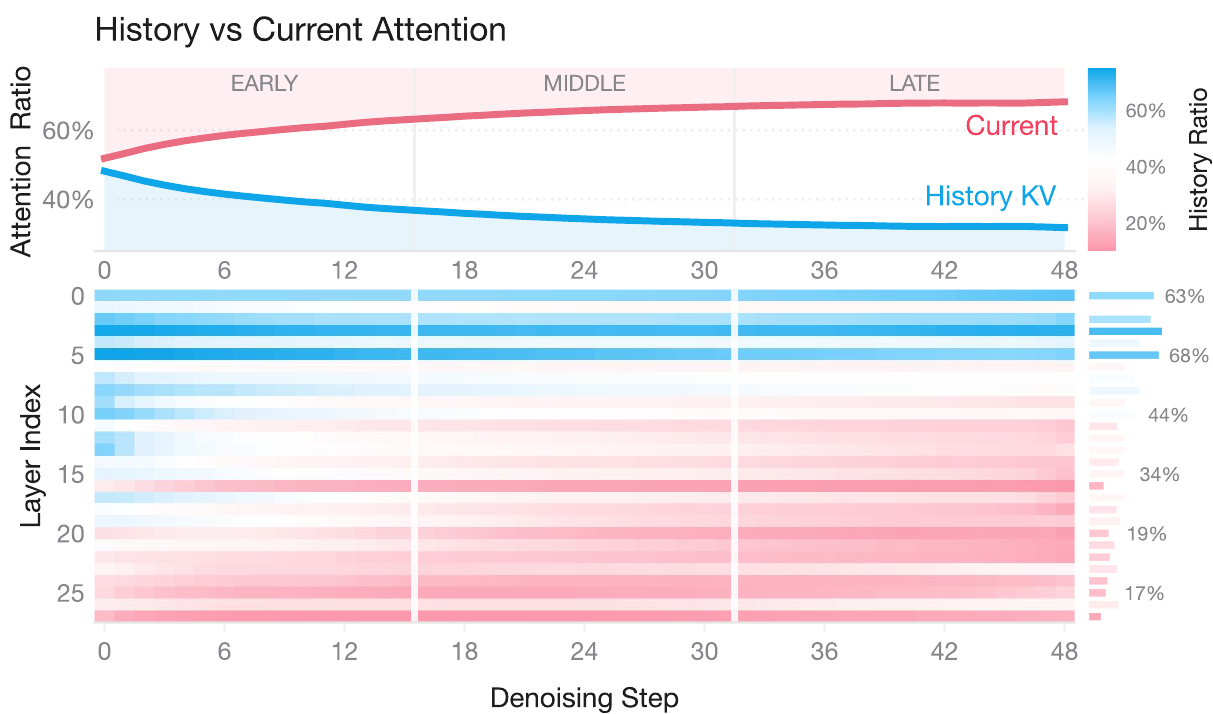}
  \caption{\textbf{Attention shifts across flow-matching steps.}
    We plot the attention mass allocated to historical-context tokens and current-generation tokens across flow-matching steps.
  This step-dependent reallocation motivates a \emph{uniform} context-selection policy across steps: step-varying visibility can cause inconsistent intermediate conditioning and destabilize generation.}
  \label{fig:stepwise_attn}
\end{figure}

Unified generators based on diffusion / flow refine the same image latents over multiple denoising steps, and the role of context can vary across this trajectory.
\Cref{fig:stepwise_attn} visualizes how attention is allocated between \emph{historical-context} tokens (history KV) and \emph{current-generation} tokens as denoising proceeds.
The top panel shows a clear step-wise shift: as denoising advances, a larger fraction of attention mass is placed on the current tokens while the share allocated to history decreases.
The bottom heatmap further suggests that this reallocation is not uniform across depth: early layers tend to devote a larger share of attention to history, whereas later layers allocate relatively more attention to current-generation tokens.

\subsection{Analysis of UniLongGen Selection Results}
\label{sec:appendix_selection_analysis}

We analyze the turns selected by UniLongGen's one-shot probing mechanism to better interpret what the model considers ``relevant'' for stable long-horizon generation.
We observe an asymmetry across modalities and depth: the early text-selection probe does not necessarily prioritize the text turns that a human would judge as the most semantically relevant to the target subject, whereas the late VAE-based image-selection probe more often surfaces visual references that align with the current synthesis target.

\paragraph{Text Turn Selection.}
For text curation, the probe at Layer~1 (VAE$\to$Text) tends to favor turns that improve \emph{grounding and controllability} of the upcoming image. Importantly, these are not always the same turns a human would select by semantic relevance alone, suggesting that early-layer ``relevance'' reflects which text provides strong, model-usable conditioning signals under a long multimodal context, rather than a pure semantic match.

\paragraph{Image Turn Selection.}
In contrast, for image curation, the VAE-based probe at $\ell_{\text{syn}}$ (VAE$\to$VAE) tends to select historical images that are directly useful for \emph{visual consistency}: it often surfaces identity-consistent subject references, as well as visually compatible scenes/viewpoints/compositions (and, when relevant, stylistically compatible references). Overall, this qualitative asymmetry is consistent with depth-wise specialization in unified generators: early layers primarily support language grounding and multimodal routing, while late VAE-dominant layers are more directly tied to synthesis dynamics, making VAE-space relevance a better proxy for selecting history that helps preserve subject identity during generation.

\subsection{Relation to Long-Context Methods}
\label{sec:relation}

UniLongGen differs fundamentally from general long-context acceleration.
While token compression (e.g., pooling) aims to \emph{fit} more information under a memory budget, UniLongGen aims to \emph{curate} information for generation quality.
Our premise is that in unified generation, retaining irrelevant dense visual history is not merely wasteful but actively harmful (via attention sinks and pollution).
Thus, we prioritize \emph{relevance-filtered sparsity} over total recall.

\subsection{Limitations}
\label{sec:appendix_limitations}
UniLongGen relies on the assumption that the one-shot attention probe accurately reflects generative needs. In cases of extremely ambiguous prompts or weak initial grounding, the relevance signal may be noisy. Additionally, the budgets ($K_{\text{grd}}, K_{\text{img}}$) are hyperparameters; while stable in our experiments, optimal values may vary across model architectures.

\section{Benchmark and Evaluation Details}
\label{sec:appendix_benchmark_eval}

\subsection{Long-Horizon Narrative Benchmark}
\label{sec:appendix_benchmark}

To rigorously evaluate the capabilities of our model in long-horizon interleaved generation, particularly focusing on narrative coherence and identity preservation, we established a specialized evaluation protocol.

\paragraph{Dataset Construction.}
Standard image generation benchmarks often focus on single-turn text-to-image synthesis, which fails to capture the complexities of maintaining context over extended sequences. To bridge this gap, we constructed a Long-Horizon Narrative Benchmark consisting of 50 curated narrative sequences. Each sequence is meticulously designed with a pre-defined textual backbone, comprising a coherent plot progression and specific descriptions for visual content.
In our main setting, each narrative contains 40 image-generation slots (i.e., $N{=}40$ in the interleaved sequence), yielding 2,000 images to be generated per full benchmark run, making it a non-trivial long-horizon stress test.

\paragraph{Interleaved Structure.}
The data is structured as an interleaved sequence:
\begin{equation}
  \mathcal{S} = \{T_1, I_1, T_2, I_2, \ldots, T_N, I_N\},
\end{equation}
where $T_i$ and $I_i$ denote the $i$-th textual and image components, respectively.

\paragraph{Task Formulation.}
We adopt a fixed-text protocol where the textual components $\{T_i\}$ serve as the condition. The model's objective is to generate the image $I_i$ given the cumulative context:
\begin{equation}
  \mathcal{C}_i = \{T_1, I_1, \ldots, T_i\}.
\end{equation}
This setup isolates the model's visual generation capabilities from text generation variance, allowing for a focused assessment of visual consistency and context adherence.

\paragraph{Subject Recurrence Strategy.}
To stress-test subject identity consistency, each narrative incorporates specific subjects that recur at various intervals. Crucially, these subjects are depicted across diverse contexts, varying in plot scenarios, background environments, character actions, and camera perspectives. This variance ensures that high consistency scores reflect robust identity learning rather than simple pixel-level memorization.

\subsection{Evaluation Metrics}
\label{sec:appendix_metrics}

We adopt a multi-axis evaluation protocol that combines (i) preference-model metrics for image quality and prompt adherence, (ii) embedding-based similarity for identity preservation, and (iii) Large Multimodal Model (LMM) judgments (we use \textbf{GPT-5.2}) for complementary, human-readable assessments of quality and long-horizon coherence.

\paragraph{Per-image quality and prompt adherence.}
We evaluate each generated image with the following automatic metrics:
\begin{itemize}[leftmargin=1.5em,itemsep=2pt]
  \item \textbf{HPS v3.} We report Human Preference Score v3 (HPSv3), a learned human-preference model trained on large-scale, diverse preference annotations. Given an image (optionally paired with its prompt), HPSv3 outputs a scalar score that is designed to correlate with human judgments of overall visual quality and preference (including aesthetics and perceived fidelity). Higher is better.
  \item \textbf{PickScore.} We report PickScore, a CLIP-based preference model trained on the Pick-a-Pic dataset of real-user pairwise preferences for text-to-image generations. PickScore takes a text prompt and an image as input and produces a scalar that reflects how likely humans would prefer the image under that prompt. In practice, it serves as a proxy for prompt adherence and perceived quality under the given text. Higher is better.
  \item \textbf{LMM judge (visual quality).} We additionally use a frontier multimodal LMM (GPT-5.2) as a reference-free judge to score each image on multiple criteria, including (a) visual fidelity and artifact level, (b) detail richness and composition, and (c) adherence to the textual description. To reduce variance, we use a fixed rubric, deterministic decoding (temperature $=0$), and aggregate scores across multiple prompts/seeds when applicable. The full prompt is presented in Figure~\ref{fig:prompt_quality}.
\end{itemize}

\paragraph{Long-horizon consistency.}
A central goal of our study is to maintain coherence over long interleaved sequences. We evaluate consistency along two complementary axes:

\begin{itemize}[leftmargin=1.5em,itemsep=2pt]
  \item \textbf{Subject identity consistency.} We measure whether a recurring subject (character/object) preserves its identity across the sequence.
    \begin{itemize}[leftmargin=1em,itemsep=1pt]
      \item \emph{Embedding-based (DINOv2).} We use DINOv2, a self-supervised vision transformer that produces general-purpose image embeddings, to quantify identity similarity. Specifically, we compute cosine similarity between the embedding of the subject crop (or region-of-interest) and a reference anchor image of the same subject, with higher similarity indicating better identity preservation.
      \item \emph{LMM-based.} We use an LMM judge (GPT-5.2) to perform reference-based comparisons (or calibrated scoring) between the current image and the anchor. The judge is instructed to focus on stable identity cues (e.g., facial/morphological traits, distinctive markings) while discounting nuisance variations such as pose, viewpoint, illumination, and background changes. The full prompt is presented in Figure~\ref{fig:prompt_consistency}. 
    \end{itemize}
  \item \textbf{Style consistency.} To assess whether the sequence maintains a coherent artistic style, we use an LMM judge (GPT-5.2) to rate cross-image stylistic coherence (e.g., color palette, rendering/brushwork, shading, and overall visual tone) across the entire sequence or across adjacent image pairs, using a standardized rubric and deterministic decoding. The full prompt is presented in Figure~\ref{fig:prompt_style_consistency}.
\end{itemize}

\section{Implementation Details}
\label{sec:appendix_implementation}

\subsection{Model and Architecture}
\label{sec:appendix_model}

We adopt Bagel as our unified multimodal backbone. Bagel is built upon a large language model (LLM) and augments it with two visual pathways: (i) a \emph{visual understanding} branch that encodes images into patch tokens using a ViT, and (ii) a \emph{visual generation} branch that represents images in a VAE latent space.

\subsection{Classifier-Free Guidance Configuration}
\label{sec:appendix_cfg}

\paragraph{Dual-branch CFG with text- and image-specific pre-contexts.}
Our inference maintains three KV-cache contexts: (i) a full context $\mathcal{C}$ used for conditional generation, (ii) a text-CFG pre-context $\mathcal{C}_{\text{text}}$ (missing the most recent generated text segment), and (iii) an image-CFG pre-context $\mathcal{C}_{\text{img}}$ (excluding any image-conditioned tokens).

\paragraph{Construction of $\mathcal{C}_{\text{text}}$ (cfg\_text).}
During prompt ingestion, before appending a text input to the full context, we snapshot the current full context as $\mathcal{C}_{\text{text}} \leftarrow \mathcal{C}$. During interleaved generation cycles, after each cycle produces a new text segment, the full context is updated with this text, while $\mathcal{C}_{\text{text}}$ is intentionally kept one-step behind (i.e., it does not include the latest generated text at the moment we call the image generator). After an image is generated and appended to the full context, we refresh $\mathcal{C}_{\text{text}} \leftarrow \mathcal{C}$ for the next cycle. Consequently, cfg\_text measures the model prediction without the current-cycle text instruction, enabling text guidance to act on the incremental effect of the latest text.

\begin{figure*}[h]
  \centering
  \includegraphics[width=\linewidth]{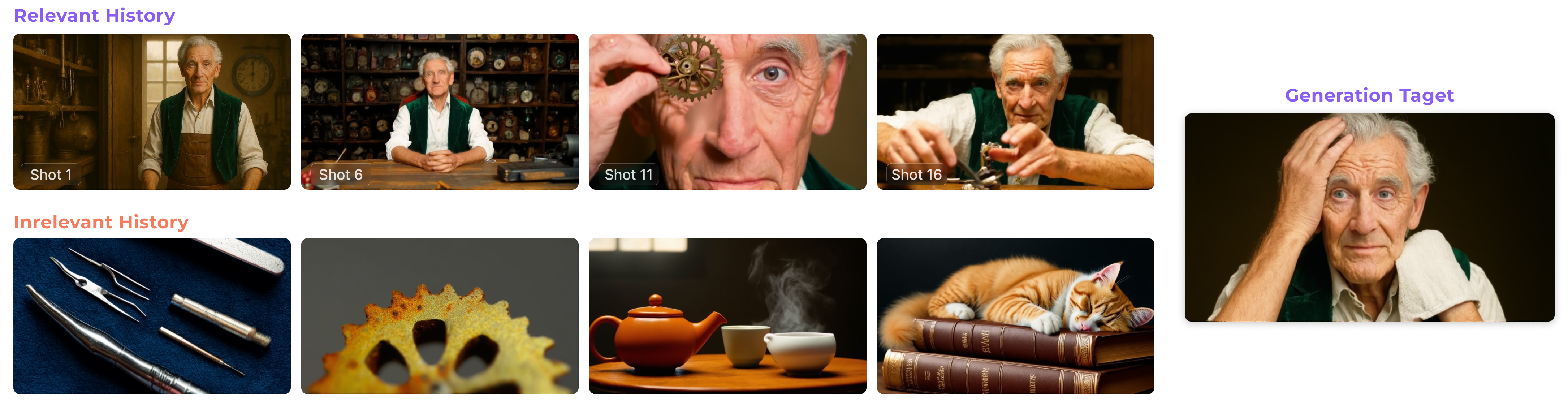}
  \caption{\textbf{Illustration of relevant vs.\ irrelevant visual history.}
    We construct sequences containing both relevant and irrelevant historical images to study attention misallocation.
    \emph{Top row (Relevant History):} Historical images depicting the same subject (an elderly clockmaker) across different shots and contexts (Shot~1, 6, 11, 16), which should receive high attention for identity-consistent generation.
    \emph{Bottom row (Irrelevant History):} Unrelated historical images (tools, gears, teapot, cat) that share no semantic connection with the generation target but may still attract spurious attention due to low-level visual similarity.
  \emph{Right (Generation Target):} The model's generation output. When irrelevant history is present, attention can be misallocated to visually salient but semantically irrelevant tokens, leading to degraded identity consistency or visual artifacts.}
  \label{fig:visual_memory_examples}
\end{figure*}

\paragraph{Construction of $\mathcal{C}_{\text{img}}$ (cfg\_img).}
We build $\mathcal{C}_{\text{img}}$ by updating it only with textual inputs from \texttt{input\_lists}. Notably, we never append any image tokens (neither input images nor generated images) to $\mathcal{C}_{\text{img}}$, and we do not update it with cycle-wise generated texts. Therefore, cfg\_img provides a stable ``text-only'' baseline that removes all image conditioning, and image guidance is computed relative to this baseline.

\paragraph{Guidance parameters and schedule.}
We use two guidance scales: \texttt{cfg\_text\_scale} (default $4.0$) and \texttt{cfg\_img\_scale} (default $1.5$). Guidance is only enabled within a diffusion-time interval \texttt{cfg\_interval} (default $[0.4, 1.0]$); outside this interval both scales are set to $1.0$ (no CFG). Sampling uses \texttt{num\_timesteps} (default $50$) with a timestep warping parameter \texttt{timestep\_shift} (default $3.0$).

\subsection{Attention Misallocation under Redundant Image History}
\label{sec:appendix_visual_memory}

This section provides experimental details and qualitative examples for the attention misallocation analysis presented in \cref{fig:visual_memory}.

\paragraph{Scene and history design.}
We construct controlled long-horizon interleaved contexts where a recurring subject appears multiple times across diverse shots (e.g., wide/medium/close-up views and different actions), forming a small set of \emph{relevant} historical images that should be informative for identity-consistent synthesis.
To stress-test attention allocation, we additionally insert \emph{irrelevant} historical images that are semantically unrelated to the target subject.
\Cref{fig:visual_memory_examples} illustrates one such setup: the top row shows four relevant subject images (Shot~1/6/11/16), while the bottom row shows unrelated distractors (e.g., tools, gears, a teapot scene, and a cat).
The generation target (right) depicts the same subject, so an ideal mechanism would concentrate attention on the relevant subject history while suppressing distractors.

\paragraph{Attention measurement.}
During the synthesis step for the target image, we aggregate attention from current-image VAE tokens (queries) to historical visual tokens (keys/values) and summarize it at the \emph{image-event} level by grouping historical tokens by image block.
We report (i) the attention mass assigned to each historical image block and (ii) the \emph{key-reference attention mass} used in the main text: the attention allocated to a known relevant reference image, normalized by the total attention over historical-image tokens.
This protocol highlights two complementary failure modes under redundant visual history: (a) non-trivial attention assigned to semantically irrelevant images (misallocation), and (b) progressive erosion of attention to true references as additional distractor images are appended (competition-induced decay).

\begin{figure*}[h]
  \centering
  \includegraphics[width=\linewidth]{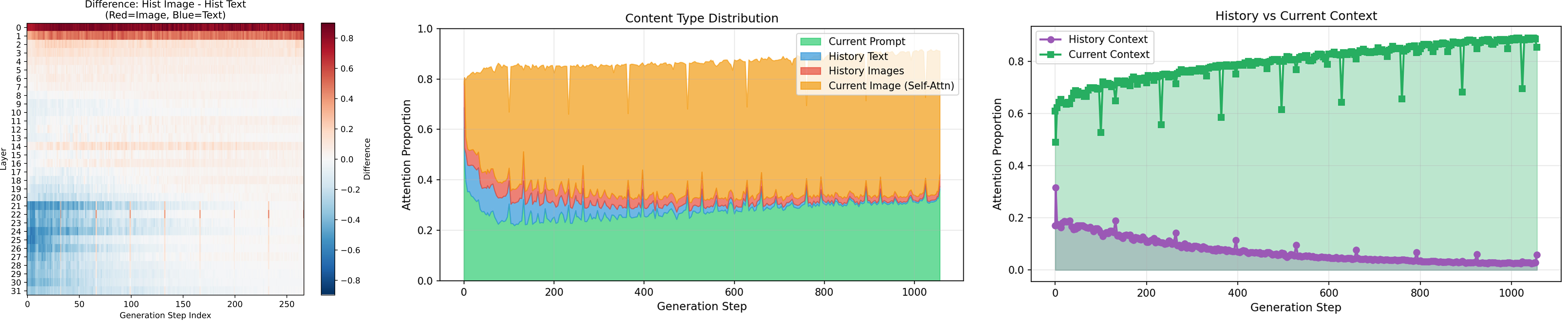}
  \caption{\textbf{Lumina-mGPT attention signature during long-horizon interleaved image decoding (pure autoregressive model).}
  \emph{Left:} layer- and step-wise difference between attention to historical \textbf{image} tokens and historical \textbf{text} tokens (Hist Image $-$ Hist Text; red indicates more attention to image history, blue indicates more attention to text history).
  \emph{Middle:} distribution of attention mass across content types over generation steps, including the current prompt, history text, history images, and current-image self-attention.
  \emph{Right:} history-context versus current-context attention proportion over generation steps.
  Overall, early layers tend to emphasize historical image content while deeper layers shift toward text, and the total attention share devoted to history is relatively small compared to current-context/self-attention.}
  \label{fig:lumima}
\end{figure*}

\section{Generalization Across Unified Models}
\label{sec:appendix_generalization}

UniLongGen is \emph{model-aligned} by design: it curates history using the model's own attention-derived relevance signals rather than an external semantic retriever.
As a result, the core principle is architecture-agnostic, but the most effective \emph{instantiation} can depend on how a given model allocates attention across layers and modalities.
Our main experiments use Bagel, a hybrid AR$+$diffusion unified model. To test whether the approach remains useful under a substantially different generation paradigm, we additionally study a pure autoregressive unified model, \textbf{Lumina-mGPT}~\cite{liu2024lumina-mgpt} (built on Chameleon~\cite{chameleon2024}), where images are generated token-by-token in the same stream.

\paragraph{Adapting the probe to the AR attention signature.}
Empirically, we find that Lumina-mGPT exhibits a different attention signature from Bagel during image generation (\cref{fig:lumima}).
In particular, the layer-wise trend can be \emph{reversed} relative to the AR$+$diffusion case: early layers may emphasize historical \emph{image} content, while deeper layers shift toward \emph{text} (left panel).
Moreover, the absolute attention share devoted to the history context can be relatively small compared to current-context/self-attention (middle/right panels), suggesting that long-horizon behavior in pure AR decoding may be less dominated by sustained history competition than in multi-step synthesis.
To respect this model-specific allocation, we keep the same model-aligned curation pipeline but adjust the probing details: we query history using the image-start special token (a stable anchor that precedes each image block) and perform image-history selection using Layer~1 attention over historical image blocks, where the model most consistently consults visual history.

\paragraph{Results and interpretation.}

As summarized in Table~\ref{tab:generalization}, this adapted UniLongGen variant improves long-horizon interleaved generation on Lumina-mGPT, but the gain over a simple sliding-window baseline is marginal.
We view this gap as informative rather than contradictory, and it suggests several practical insights about \emph{when} model-aligned curation matters most.

(1) First, the \textbf{failure mode differs across generation paradigms}. In hybrid AR$+$diffusion models, each image is synthesized through repeated refinement steps where dense visual history can repeatedly enter the conditioning pathway; this creates sustained softmax competition and makes ``which events remain visible'' particularly consequential. In contrast, in pure AR image decoding, generation proceeds token-by-token and the model's attention may be more dominated by current-context/self-attention (\cref{fig:lumima}), so recency truncation can already remove many competitors at low risk, making sliding windows a stronger baseline.

(2) Second, the relative gain depends on the model's \textbf{capacity and training coverage for interleaved generation}. Lumina-mGPT's multi-image generation capability is substantially weaker than Bagel's: it produces virtually no cross-image identity or style consistency even in short sequences, indicating that the model has not internalized the ability to leverage historical visual context for coherent synthesis. Because of this near-absence of baseline consistency, we omit identity and style consistency metrics for Lumina-mGPT altogether---the scores are too low to be meaningfully discriminative. In this regime, long-horizon degradation is dominated by the model's intrinsic capability gap rather than an attention-allocation bottleneck, leaving any history-selection policy with minimal headroom beyond what simple truncation already provides.

(3) Third, the Lumina-mGPT case highlights a \textbf{general recipe for cross-model transfer}: UniLongGen's mechanism is not tied to a fixed layer split or a fixed query token, but to aligning selection with the model's own attention signature. In practice, one can diagnose this signature by measuring layer-wise modality ratios during image generation, then choose probing layers and query anchors that correspond to where the model actually consults (or should consult) historical images versus text.

Overall, these results suggest that the largest benefits of model-aligned event-level curation arise when (i) dense visual history actively competes for attention throughout synthesis and (ii) the model is capable of leveraging historical references to maintain long-range consistency.

\begin{table}[h]
  \centering
  \small
  \setlength{\tabcolsep}{4pt}
  \renewcommand{\arraystretch}{1.15}
  \caption{\textbf{Generalization to a pure autoregressive unified model (Lumina-mGPT).}}
  \label{tab:generalization}
  \vspace{2pt}
  \resizebox{\columnwidth}{!}{%
  \begin{tabular}{@{}llcccc@{}}
    \toprule
    \textbf{Model} & \textbf{Method} & \textbf{HPS v3}$\uparrow$ & \textbf{Qual.}$\uparrow$  \\
    \midrule
    \multirow{2}{*}{Lumina-mGPT (Chameleon-based)} 
    & Dense KV & 2.7954 & 0.1904 \\
    & Sliding Window & 7.2965 & 0.2014 \\
    & + UniLongGen & 7.3020 & 0.2015 \\
    \bottomrule
  \end{tabular}
  }%
\end{table}

\begin{figure*}[t]
  \begin{center}
    \centerline{\includegraphics[width=\linewidth]{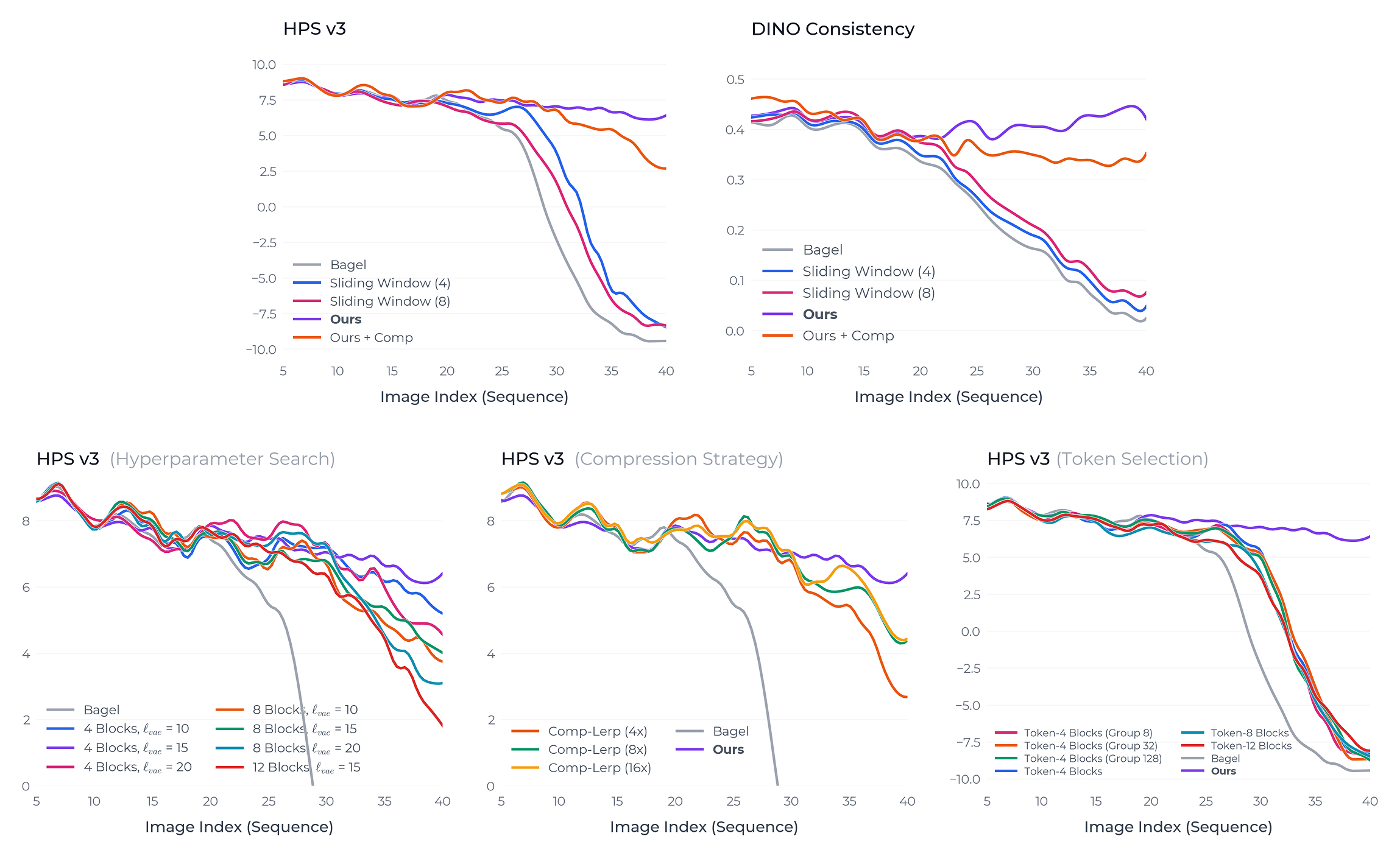}}
    \caption{
      \textbf{Position-wise performance curves for long-horizon interleaved generation.}
      \emph{Top:} HPS v3 (quality) and DINO consistency (identity) as a function of image index in a 40-image sequence, comparing the base model (Bagel), anchored sliding-window baselines (keep first + last $N$ images), UniLongGen, and UniLongGen with compression.
      \emph{Bottom:} HPS v3 ablations for (left) hyperparameters (number of retained image blocks and VAE probing layer), (middle) compression strategies (LERP with different downsampling rates), and (right) token-level selection variants, showing that UniLongGen best preserves long-horizon stability.
    }
    \label{fig:curve}
  \end{center}
\end{figure*}

\section{Additional Results}

We provide additional qualitative examples of long-horizon interleaved text--image generation in \cref{fig:supp_results_1,fig:supp_results_2,fig:supp_results_3,fig:supp_results_4,fig:supp_results_5,fig:supp_results_6,fig:supp_results_7}.
Each example is a continuous narrative with \textbf{40} image slots generated autoregressively in an interleaved sequence, illustrating long-range fidelity and identity/style consistency over many image events.

\paragraph{Position-wise curves.}
\Cref{fig:curve} provides position-wise performance curves on the 40-image long-horizon benchmark, complementing the aggregate metrics in the main ablation tables.
The top row plots HPS v3 (quality) and DINO consistency (identity) versus image index, contrasting the base model (dense KV), anchored sliding-window baselines (keep the first anchor plus the last $N$ images), and UniLongGen variants.
The bottom row further visualizes how key design choices affect long-horizon stability, including hyperparameter choices (retention budget and probing layer), discard handling via compression, and token-level selection variants.
Overall, these curves qualitatively mirror the conclusions drawn from the summary metrics (e.g., \cref{tab:big_ablation,tab:vae_layer_ablation}): long-horizon degradation is position-dependent, and UniLongGen's event-level curation most consistently delays or mitigates collapse across the sequence.

\section{Additional Ablations}
\label{sec:appendix_ablations}

\begin{figure}[h]
  \begin{center}
    \centerline{\includegraphics[width=\columnwidth]{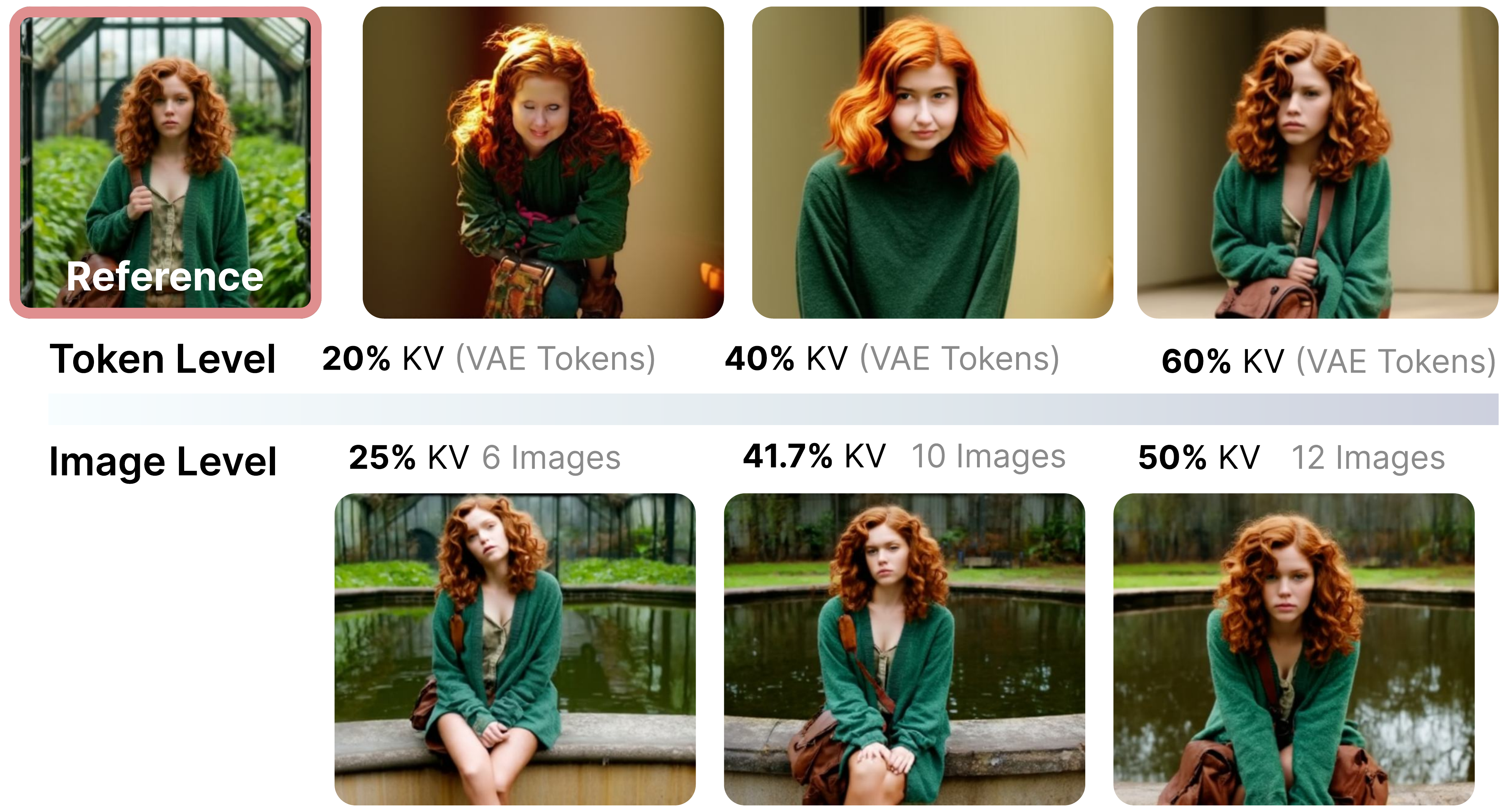}}
    \caption{
      \textbf{Token-level vs.\ image-level history selection under matched KV budgets.}
      We compare two ways to reduce the visible KV cache when conditioning on long visual history.
      \emph{Top (token level):} keep a fixed fraction of historical VAE tokens (e.g., 20\%/40\%/60\% KV), which can weaken the effective reference signal and lead to degraded subject fidelity in this example.
      \emph{Bottom (image level):} keep a small number of whole reference images at comparable KV budgets (e.g., 6/10/12 images), which better preserves identity-consistent conditioning.
      This qualitative comparison supports event/image-level curation over token-level pruning for stable long-horizon synthesis.
    }
    \label{fig:token_budget}
  \end{center}
\end{figure}

\subsection{Token-level vs.\ image-level selection.}
\Cref{fig:token_budget} provides a qualitative illustration of the selection-granularity effect discussed in \cref{sec:experiments}.
Under comparable KV budgets, token-level pruning of historical VAE tokens can remove or fragment critical visual cues (while leaving enough dense keys to still participate in softmax competition), resulting in weaker identity-preserving conditioning.
In contrast, retaining a small set of whole reference images concentrates the budget on a few coherent events, which is often more effective for maintaining subject consistency in long interleaved generation.

\begin{table}[h]
  \centering
  \small
  \setlength{\tabcolsep}{5.5pt}
  \renewcommand{\arraystretch}{1.15}
    \caption{\textbf{Ablation: using Post-Softmax attention vs.\ Pre-Softmax $QK$ scores for relevance evaluation.}
    We compare two ways to compute the relevance signal used to select historical reference images: (i) aggregating \emph{Post-Softmax} attention weights, and (ii) using \emph{Pre-Softmax} head-averaged $QK$ similarity scores.
    We find that using the Pre-Softmax score yields substantially better long-horizon generation quality and identity consistency.}
  \label{tab:head_agg_ablation}
  \resizebox{\columnwidth}{!}{%
    \begin{tabular}{@{}lcccc@{}}
      \toprule
      \textbf{Variant} & \textbf{HPS v3}$\uparrow$ & \textbf{PickScore}$\uparrow$ & \textbf{Style Cons.}$\uparrow$ & \textbf{DINOv2-ID}$\uparrow$ \\
      \midrule
      Baseline  & 3.1677 & 0.1943 & 3.99 & 0.3164 \\
      Post-Softmax  & 5.2779 & 0.1978 & 5.44 & 0.3617 \\
      \rowcolor{oursRow}
      \textbf{Pre-Softmax (Ours)} & \best{7.5701} & \best{0.2025} & \best{6.13} & \best{0.4272} \\
      \bottomrule
    \end{tabular}
  }%
\end{table}

\subsection{Pre-Softmax vs.\ Post-Softmax}
\label{sec:exp_head_agg}
We additionally ablate how to compute the \emph{relevance signal} for selecting historical reference images: using (i) \emph{Post-Softmax} attention weights, or (ii) \emph{Pre-Softmax} $QK$ similarity scores (no softmax).
Concretely, Post-Softmax first computes per-head attention (including softmax normalization) and then aggregates across heads, whereas Pre-Softmax first aggregates current-image VAE queries (per head) and then computes head-averaged $QK$ similarity as the relevance score.
\textbf{Why Post-Softmax tends to be biased toward recency.}
After softmax, attention is typically top-heavy and dominated by a few highest-scoring keys; in long histories, the most recent turns often contain tokens that are easiest to match (and are less degraded by intervening context), so they frequently receive the largest attention mass.
As a result, Post-Softmax relevance estimation can collapse to a near-recency heuristic, repeatedly selecting the most recent few image blocks.
As shown in Table~\ref{tab:head_agg_ablation}, using Pre-Softmax $QK$ scores yields substantially better long-horizon quality and identity/style consistency and mitigates this recency bias.

\definecolor{promptgreen}{RGB}{0, 128, 0}
\definecolor{promptbg}{RGB}{245, 255, 245}

\begin{figure*}[h!]
  \centering
  \begin{tcolorbox}[
      enhanced,
      width=\textwidth,
      colback=promptbg,
      colframe=promptgreen,
      coltitle=promptgreen,
      fonttitle=\bfseries\large,
      title={Prompt C.1 (Subject Consistency)},
      attach boxed title to top left={xshift=6mm, yshift*=-\tcboxedtitleheight/2},
      boxed title style={
        colback=white,
        colframe=promptgreen,
        boxrule=1pt,
        arc=2pt,
        top=3pt, bottom=3pt, left=5pt, right=5pt
      },
      arc=2pt,
      boxrule=1pt,
      left=6mm, right=6mm, top=6mm, bottom=4mm,
      toptitle=2mm
    ]
    \small

    \textbf{Task Description}
    You are an expert at evaluating character identity consistency in AI-generated images. Your task is to compare the \textbf{TARGET IMAGE} with the \textbf{REFERENCE IMAGE} and evaluate how well the \textbf{SAME CHARACTER's} identity is preserved.

    \vspace{0.5em}
    \textbf{Evaluation Criteria}
    \begin{itemize}[leftmargin=1.5em, nosep]
      \item \textbf{Focus ONLY on Character Identity}: Facial features (shape, eyes, nose, mouth), Hair (color, style), Accessories, Body type, and Skin tone.
      \item \textbf{MUST IGNORE}: Lighting, Color grading, Camera angle, Shot type, Pose/Expression, Background, Clothing (unless uniform), Image quality.
    \end{itemize}

    \vspace{0.5em}
    \textbf{Scoring Guidelines}
    Score the identity consistency on a scale of 1.0 to 10.0. Be discriminative!
    \begin{itemize}[leftmargin=1.5em, itemsep=2pt]
      \item \textbf{1.0--2.0}: Completely different person, unrecognizable.
      \item \textbf{2.1--4.0}: Major identity changes, barely recognizable as same character.
      \item \textbf{4.1--5.5}: Noticeable identity inconsistencies, some features changed.
      \item \textbf{5.6--7.0}: Mostly consistent, minor variations in some features.
      \item \textbf{7.1--8.5}: Highly consistent, same person clearly recognizable.
      \item \textbf{8.6--10.0}: Perfect identity preservation, identical character appearance.
    \end{itemize}

    \vspace{0.5em}
    \textbf{Output Format}
    Respond ONLY with valid JSON:
    \begin{tcolorbox}[colback=white, colframe=promptgreen!30, boxrule=0.5pt, arc=0pt, left=2mm, top=2mm, bottom=2mm]
      \ttfamily
      \{"analysis": "Brief analysis focusing on identity consistency (under 40 words), "score": X.X"\}
    \end{tcolorbox}

  \end{tcolorbox}
  \caption{The prompt used for GPT-5.2 based subject consistency evaluation.}
  \label{fig:prompt_consistency}
\end{figure*}

\begin{figure*}[h!]
  \centering
  \begin{tcolorbox}[
      enhanced,
      width=\textwidth,
      colback=promptbg,
      colframe=promptgreen,
      coltitle=promptgreen,
      fonttitle=\bfseries\large,
      title={Prompt C.2 (Style Consistency)},
      attach boxed title to top left={xshift=6mm, yshift*=-\tcboxedtitleheight/2},
      boxed title style={
        colback=white,
        colframe=promptgreen,
        boxrule=1pt,
        arc=2pt,
        top=3pt, bottom=3pt, left=5pt, right=5pt
      },
      arc=2pt,
      boxrule=1pt,
      left=6mm, right=6mm, top=6mm, bottom=4mm,
      toptitle=2mm
    ]
    \small

    \textbf{Task Description}
    You are an expert art critic and visual style analyst. Your task is to evaluate the \textbf{ARTISTIC STYLE CONSISTENCY} across a series of AI-generated images. Analyze the provided images and evaluate how consistent their artistic style is across all frames.

    \vspace{0.5em}
    \textbf{Evaluation Criteria}
    \begin{itemize}[leftmargin=1.5em, nosep]
      \item \textbf{Focus ONLY on Artistic Style}: Rendering style (realistic, stylized, etc.), Color palette usage, Brushwork/Texture, Shading technique, Line work, and Overall aesthetic.
      \item \textbf{MUST IGNORE}: Subject matter, Lighting conditions, Color temperature, Camera angle, Characters/Objects, Scene/Environment, Time of day, Image quality (artifacts, blur), and Generation defects.
    \end{itemize}
    \emph{Note: Style consistency is about whether images look like they came from the same artist/model, NOT whether each image is well-generated.}

    \vspace{0.5em}
    \textbf{Scoring Guidelines}
    Score the style consistency on a scale of 1.0 to 10.0. Be discriminative!
    \begin{itemize}[leftmargin=1.5em, itemsep=2pt]
      \item \textbf{1.0--2.0}: Completely inconsistent styles, looks like different artists/models.
      \item \textbf{2.1--4.0}: Major style inconsistencies, some images clearly different.
      \item \textbf{4.1--5.5}: Moderate inconsistency, noticeable style variations.
      \item \textbf{5.6--7.0}: Mostly consistent, minor style variations.
      \item \textbf{7.1--8.5}: Highly consistent style, cohesive visual language.
      \item \textbf{8.6--10.0}: Perfect style consistency, appears from same artist/model.
    \end{itemize}

    \vspace{0.5em}
    \textbf{Output Format}
    IMPORTANT: First analyze the style elements, THEN determine the score. Respond ONLY with valid JSON:
    \begin{tcolorbox}[colback=white, colframe=promptgreen!30, boxrule=0.5pt, arc=0pt, left=2mm, top=2mm, bottom=2mm]
      \ttfamily
      \{
        "analysis": "xxx (60-80 words)",
        "score": X.X
      \}
    \end{tcolorbox}

  \end{tcolorbox}
  \caption{The prompt used for GPT-5.2 based artistic style consistency evaluation.}
  \label{fig:prompt_style_consistency}
\end{figure*}

\begin{figure*}[h!]
  \centering
  \begin{tcolorbox}[
      enhanced,
      width=\textwidth,
      colback=promptbg,
      colframe=promptgreen,
      coltitle=promptgreen,
      fonttitle=\bfseries\large,
      title={Prompt C.3 (Visual Quality)},
      attach boxed title to top left={xshift=6mm, yshift*=-\tcboxedtitleheight/2},
      boxed title style={
        colback=white,
        colframe=promptgreen,
        boxrule=1pt,
        arc=2pt,
        top=3pt, bottom=3pt, left=5pt, right=5pt
      },
      arc=2pt,
      boxrule=1pt,
      left=6mm, right=6mm, top=6mm, bottom=4mm,
      toptitle=2mm
    ]
    \small

    \textbf{Task Description}
    You are an expert evaluator specializing in AI-generated image quality assessment. Your task is to evaluate the visual quality of the generated image.

    \vspace{0.5em}
    \textbf{Evaluation Criteria}
    Focus on the following aspects:
    \begin{itemize}[leftmargin=1.5em, nosep]
      \item Sharpness and clarity: Is the image crisp and well-defined?
      \item Detail richness: Are fine details rendered with precision?
      \item Color harmony: Are colors balanced, natural, and aesthetically pleasing?
      \item Composition: Is the layout visually balanced and well-structured?
      \item Artifact detection: Are there typical AI generation flaws (deformations, anatomical errors, unnatural textures, blending issues, repetitive patterns)?
      \item Overall aesthetic quality: Does it look professional and visually appealing?
    \end{itemize}

    \vspace{0.5em}
    \textbf{Scoring Guidelines (Provide discriminative scores)}
    \begin{itemize}[leftmargin=1.5em, itemsep=2pt]
      \item \textbf{1.0--2.0}: Severe artifacts, blurry, distorted, unrecognizable
      \item \textbf{2.1--4.0}: Obvious defects, multiple generation errors visible
      \item \textbf{4.1--6.0}: Average quality, noticeable issues present
      \item \textbf{6.1--7.5}: Good quality, minor imperfections
      \item \textbf{7.6--9.0}: Excellent quality, refined details, visually appealing
      \item \textbf{9.1--10.0}: Professional-grade quality, virtually flawless
    \end{itemize}

    \vspace{0.5em}
    \textbf{Output Format}
    Respond ONLY with valid JSON, no additional text:
    \begin{tcolorbox}[colback=white, colframe=promptgreen!30, boxrule=0.5pt, arc=0pt, left=2mm, top=2mm, bottom=2mm]
      \ttfamily
      \{"analysis": "Brief analysis (under 50 words)", "score": X.X\}
    \end{tcolorbox}

  \end{tcolorbox}
  \caption{The prompt used for GPT-5.2 based visual quality evaluation.}
  \label{fig:prompt_quality}
\end{figure*}

\end{document}